\documentclass[twoside]{article}

\usepackage[accepted]{aistats2026}
\usepackage{amsmath,amsfonts,bm}

\def\eqref#1{equation~\ref{#1}}

\def\1{\bm{1}}

\DeclareMathAlphabet{\mathsfit}{\encodingdefault}{\sfdefault}{m}{sl}
\SetMathAlphabet{\mathsfit}{bold}{\encodingdefault}{\sfdefault}{bx}{n}

\usepackage[linesnumbered,ruled,vlined]{algorithm2e}
\usepackage{setspace}
\usepackage{amsfonts}       
\usepackage{booktabs}       
\usepackage{enumitem}
\usepackage[T1]{fontenc}    
\usepackage{hyperref}       
\usepackage[utf8]{inputenc} 
\usepackage{makecell}
\usepackage{microtype}      
\usepackage{multirow}
\usepackage{nicefrac}       
\usepackage{tabularx}
\usepackage{url}            
\usepackage{xcolor}         
\usepackage{wrapfig}
\usepackage{xr}
\usepackage[round]{natbib}

\usepackage{minitoc}
\usepackage{float}
\usepackage{mathrsfs}
\usepackage{csquotes}
\usepackage{subfigure}
\usepackage{etoolbox}

\usepackage{filecontents,catchfile,pgffor}

\usepackage{environ}
\usepackage[percent]{overpic}
\usepackage[dvipsnames]{xcolor}

\usepackage{amsmath}
\usepackage{amssymb}
\usepackage{algpseudocode}
\usepackage{booktabs}
\usepackage{dsfont}
\usepackage{graphicx}
\usepackage{makecell}
\usepackage{multirow}
\usepackage{pifont}
\usepackage[group-separator={,}]{siunitx}
\usepackage{tabularx}
\usepackage{caption}
\usepackage{soul}
\usepackage{changepage,threeparttable}
\usepackage{colortbl}
\usepackage[accsupp]{axessibility}  
\usepackage{mathtools}
\usepackage{xfrac}
\usepackage{enumitem}
\usepackage{catchfile}
\usepackage{longtable}
\usepackage{tabu}
\usepackage{tcolorbox}
\usepackage{supertabular}
\usepackage{caption}
\usepackage{subcaption}
\usepackage{pifont}
\usepackage{adjustbox}
\usepackage{titletoc}
\usepackage{multicol}
\newcolumntype{Y}{>{\centering\arraybackslash}X}
\usepackage{physics}

\usepackage{amsthm}
\newtheorem{theorem}{Theorem}
\newtheorem{proposition}{Proposition}
\newtheorem*{proposition*}{Proposition}
\newtheorem{lemma}{Lemma}
\newtheorem{corollary}{Corollary}
\newtheorem*{corollary*}{Corollary}

\newcommand{\B}{\mathbf}

\renewcommand{\eqref}[1]{(\ref{#1})}

\newcommand{\ie}{i.e.,}
\newcommand{\eg}{e.g.,}

\definecolor{color_1}{RGB}{255,0,128}
\definecolor{color_2}{RGB}{128,128,0}
\definecolor{color_3}{RGB}{0,128,0}
\definecolor{color_4}{RGB}{128,0,0}
\definecolor{color_5}{RGB}{128,0,128}

\definecolor{Goldenrod}{RGB}{218, 165, 32}
\definecolor{ForestGreen}{RGB}{34, 139, 34}

\definecolor{highlight_color}{RGB}{255,0,128}
\definecolor{blue_highlight_color}{RGB}{0,128,255}
\definecolor{green_highlight_color}{RGB}{51,145,40}
\definecolor{orange_highlight_color}{RGB}{255,160,0}
\definecolor{purple_highlight_color}{RGB}{128,0,128}

\newcommand{\mycomment}[1]{}

\newif\ifpaper
\paperfalse

\begin{document}

%
\runningtitle{Demystifying Transition Matching}

%
\runningauthor{Jaihoon Kim, Rajarshi Saha, Minhyuk Sung, Youngsuk Park}

\twocolumn[

\aistatstitle{Demystifying Transition Matching: \\ When and Why It Can Beat Flow Matching}

\vspace{-1.0em}

\aistatsauthor{ 
Jaihoon Kim\textsuperscript{\textdagger, 1} \And
Rajarshi Saha\textsuperscript{\textdaggerdbl, 2} \And
Minhyuk Sung\textsuperscript{1} 
\And Youngsuk Park\textsuperscript{2} 
}

\vspace{0.25em}
\aistatsaddress{%
\textsuperscript{1}KAIST, 
\textsuperscript{2}Amazon Web Services
}

\vspace{-2.0em}
\aistatsaddress{%
    \texttt{\{jh27kim, mhsung\}@kaist.ac.kr}, 
    \texttt{\{sahrajar, pyoungsu\}@amazon.com}
}

]

\begingroup
  \def\thefootnote{\textdagger}
  \footnotetext{Work done during internship at Amazon Web Services.}
  \def\thefootnote{\textdaggerdbl}
  \footnotetext{Correspondence to Rajarshi Saha.}
\endgroup
\setcounter{footnote}{0}

\vspace{-0.5\baselineskip}
\begin{abstract}
\vspace{-0.5\baselineskip}
Flow Matching (FM) underpins many state-of-the-art generative models, yet recent results indicate that Transition Matching (TM) can achieve higher quality with fewer sampling steps. 
This work answers the question of \emph{when} and \emph{why} TM outperforms FM. 
First, when the target is a unimodal Gaussian distribution, we prove that TM attains strictly lower KL divergence than FM for finite number of steps. 
The improvement arises from stochastic difference latent updates in TM, which preserve target covariance that deterministic FM underestimates. 
We then characterize convergence rates, showing that TM achieves faster convergence than FM under a fixed compute budget. 
Second, we extend the analysis to Gaussian mixtures and identify local–unimodality regimes in which the sampling dynamics approximate the unimodal case, where TM can outperform FM. 
The approximation error decreases as the minimal distance between component means increases, highlighting that TM is favored when the modes are well separated. 
However, when the target variance approaches zero, each TM update converges to the FM update, and the performance advantage of TM diminishes. 
In summary, we show that TM outperforms FM when the target distribution has well-separated modes and non-negligible variances.
We validate our theoretical results with controlled experiments on Gaussian distributions, and extend the comparison to real-world applications in image and video generation. 
\end{abstract}
\section{INTRODUCTION}
\vspace{-0.4em}
\label{sec:intro}

Generative modeling has achieved remarkable success in image and video generation \citep{chen2025goku, kong2024:hunyuan, BlackForestLabs:2024Flux, Esser2403:scaling, wan2025:wan}, 3D content creation \citep{xiang2025:trellis, li2025:triposg}, and scientific applications such as protein and material design \citep{miller2024:flowmm, geffner2025:proteina}. 
Among these approaches, \emph{Flow Matching} (FM) \citep{Liu:2023RF, Lipman:2023CFM} has emerged as a core framework and underpins many state-of-the-art generative models \citep{BlackForestLabs:2024Flux, kong2024:hunyuan}. 
FM learns a continuous-time velocity field and generates samples by numerically simulating the corresponding ordinary differential equation (ODE) with $N$ discretization steps.
However, it is computationally demanding because each step requires a full backbone evaluation.


{\it Transition Matching} (TM) \citep{shaul2025:transition} was recently introduced as an alternative that can surpass FM in the low-step regime.
Rather than modeling a velocity field, TM learns a discrete-time transition kernel via a stochastic {\it difference latent} that specifies per-step updates.
At inference, backbone features are cached once per $N$ outer steps, while a lightweight head solves an $S$-step inner ODE for the difference latent.
Although TM converges to FM as $N \to \infty$, it is often superior when $N$ is small, and this advantage still remains largely empirical with limited theoretical understanding of when and why TM outperforms FM.

In this work, we address this gap by providing theoretical analyses clarifying TM's behavior relative to FM.
In the analytically tractable unimodal Gaussian setting, we show that for any finite $N > 1$ and $S > 1$, TM achieves strictly lower KL divergence than FM, as TM’s {\it stochastic sampling of the difference latent preserves the target covariance} that FM underestimates. 
We further establish distinct convergence rates: $\mathcal{O}(1/N^2)$ for FM vs. $\mathcal{O}(1/N^2 S^2)$ for TM.
Since increasing $N$ requires repeated backbone passes while increasing $S$ only uses a lightweight head, TM consistently attains lower KL divergence under equal compute, explaining its clear advantage in the low-step regime. 


We then extend our analysis to mixture of Gaussians, where the target distribution has $K$ components with means \(\{\mu_k\}\) and covariances \(\{\sigma_k^2 I_d\}\). 
We show that when the components are {\it sufficiently well-separated}, sampling effectively reduces to the unimodal case, with approximation error decaying exponentially in the squared minimum separation \(D_{\min}=\min_{j\neq k}\|\mu_j-\mu_k\|\). 
In this regime, TM again outperforms FM at finite steps.
However, as variances shrink, the difference latent distribution collapses to its mean, making TM’s updates indistinguishable from FM’s.
This yields a unified view: TM excels when mixture components are {\it well-separated} and possess {\it non-negligible variance}.

Finally, we complement our theoretical results with carefully designed experiments that validate the predictions in both unimodal and mixture settings. 
Beyond synthetic benchmarks, we further evaluate the framework on real-world datasets, including image generation and {\it for the first time}, video generation. 
These experiments highlight not only the tight agreement between theory and practice, but also clearly demonstrate that TM consistently outperforms FM under comparable or even lower compute budgets, reinforcing its advantage in practical generative modeling scenarios.

Overall, our contributions are summarized as follows:
\vspace{-0.5em}
\begin{itemize}[leftmargin=*]
    \item \textbf{Unimodal Gaussian: Why TM is Better.} 
    We prove that TM achieves strictly lower KL divergence than FM for finite $N>1$ and $S>1$, can converge faster under fixed compute budget, i.e., $\mathcal{O}(1/N^2)$ vs. $\mathcal{O}(1/S^2 N^2)$ (\S \ref{sec:gauss_target_main}; Thm. \ref{theorem:kl}). 
    \item \textbf{Gaussian mixtures: When TM is better.} 
    We show that well-separated components reduce to the unimodal case with error decaying exponentially in $D_{\rm min}$, and TM excels FM when modes are separated and variances are non-negligible (\S \ref{sec:multimodal_main}; Thm. \ref{thm:gaussian_mixture_KL_convergence}). 
    \item \textbf{Validation with real-world applications.} 
    We validate our theoretical insights on two large-scale generative modeling tasks: class-conditioned image generation and frame-conditioned video generation.
    For images, TM achieves superior quality–compute trade-offs, consistently surpassing FM at a lower latency.
    For video, TM, evaluated against the strong History-Guided Diffusion baseline, consistently improves upon multiple goodness-of-fit metrics under matched compute.
    This marks the first application of TM to video generation, where it both enhances temporal coherence and reduces inference cost (\S \ref{sec:application}).
\end{itemize}

Together, these results confirm that the efficiency gains predicted in unimodal and mixture analyses extend directly to challenging real-world domains, establishing TM as a practical and scalable alternative to FM in compute-constrained generative modeling.

\section{OVERVIEW OF MODELING FRAMEWORKS}
\label{subsec:background}

The goal of generative modeling is to construct a process that transports samples from a simple \emph{source distribution} \(p_0\) to samples from a \emph{target distribution} \(p_1\). 
A common choice for \(p_0\) is the standard Gaussian distribution, \(p_0=\mathcal N(0,I_d)\), due to its tractability and ease of sampling, while \(p_1\) corresponds to the empirical data distribution.
This process is formalized as a family of random variables \(\{X_t\}_{t\in[0,1]}\), such that \(X_0\sim p_0\) and \(X_1\sim p_1\).
Different families of generative models realize this in different ways.
For instance, diffusion models describe the transformation through a stochastic forward process and its corresponding reverse dynamics, while more recent flow-matching approaches learn an explicit invertible mapping.
In this work, we study Transition Matching~\citep{shaul2025:transition}, a recently proposed discrete-time, continuous-state framework that generalizes flow-matching by learning explicit transition kernels rather than deterministic mappings. 

\subsection{Flow Matching (FM)}
\label{subsec:cfm_background}
Flow models \citep{Lipman:2023CFM} describe the continuous evolution of a random variable \( \{X_t\}_{t \in [0,1]} \) from an initial distribution \( X_0 \sim p_0 \) to a target distribution \( X_1 \sim p_1 \), governed by the ODE,
\begin{equation}
    \label{eq:fm-ode}
    \frac{d}{dt} X_t = u_t(X_t), \qquad X_0 \sim p_0,
\end{equation}
where \( u_t(x): [0,1] \times \mathbb{R}^d \to \mathbb{R}^d \) is a time-dependent \emph{velocity field}. 
The solution of this ODE is referred to as a flow \( \psi_t : \mathbb{R}^d \to \mathbb{R}^d \), with \( X_t = \psi_t(X_0) \) evolving deterministically over time.
The flow induces a time-indexed family of pushforward measures, mapping $p_0$ to intermediate densities $p_t$, thereby transporting the initial distribution along a continuous path of probability densities \( \{p_t\}_{t \in [0,1]} \). 
Therefore, the objective of flow models is to learn the velocity field \( u_t(x) \) (equivalently, the corresponding flow) that induces the desired probability path between source \( p_0 \) and target \( p_1 \).

To guide this learning, flow models use the Conditional Optimal Transport (CondOT) path,
\begin{align}
    \label{eq:cond_ot}
    X_t = (1 - t) X_0 + t X_1 \sim p_t, 
\end{align}
which linearly interpolates between $X_0 \sim \mathcal{N}(0, I_d)$ and $X_1 \sim p_1$, providing a reference path for training \( u_t (x)\).

In FM, given training data consisting of samples drawn from $p_1$, a neural network $v_t^\theta$ learns the velocity field $u_t(x)$ by minimizing the following loss
\begin{equation} 
    \label{eq:fm-loss} 
    \mathcal L_{\mathrm{FM}}(\theta) = \mathbb E_{\substack{t \sim \mathcal U[0,1], \\ X_t \sim p_t}} \big\|v_t^\theta(X_t) - u_t(X_t)\big\|^2,
\end{equation}
where $\mathcal{U}[0,1]$ is the uniform distribution.
However, $u_t(x)$ in the expression above is generally untractable. 
Consequently, Conditional Flow Matching (CFM) objective reformulates \eqref{eq:fm-loss} for the conditional probability paths $p_{t|1}(x|x_1) = \mathcal{N}(t X_1, (1-t)^2 I_d)$, defined as  
\begin{align} 
    \label{eq:cfm-loss} 
    \mathcal L_{\mathrm{CFM}}(\theta) &= \mathbb E_{
    \substack{
    X_t \sim p_{t|1}, \\ X_1 \sim p_1, \\
    t \sim \mathcal U[0,1]}} \big\|v_t^\theta(X_t) - u_t(X_t | X_1) \big\|^2,
\end{align}
where \(u_t(X_t \mid X_1) = X_1 - X_0\) is the conditional velocity along the CondOT path \eqref{eq:cond_ot}. 
Notably, Thm. 2 of \citeauthor{Lipman:2023CFM} states that minimizing $\mathcal L_{\mathrm{CFM}}(\theta)$ and $\mathcal L_{\mathrm{FM}}(\theta)$ are equivalent.  
As the conditional expectation minimizes the squared loss, minimizing \(\mathcal{L}_{\rm CFM}(\theta)\) yields
\begin{equation}
    \label{eq:FM_expected_velocity_update}
    v_t^{\theta}(X_t) = \mathbb{E}_{X_0, X_1}[X_1 - X_0 \vert X_t].
\end{equation}
Following convention, we refer to the models trained with CFM loss \eqref{eq:cfm-loss} as flow models (FM). 
At inference time, data is sampled by simulating the ODE in \eqref{eq:fm-ode} (\eg~Euler method) using the learned $v_t^{\theta}$. 
Given a total number of steps $N$, the interval $[0,1]$ is discretized as $t_n = n\Delta t$, where $\Delta t = 1/N$ and $n \in [N-1] \triangleq \{0, \ldots, N-1\}$. 
Each Euler step proceeds as follows:
\begin{equation}
    \label{eq:fm-euler}
    \widetilde X_{t_{n+1}}^{\mathrm{FM}} = \widetilde X_{t_n}^{\mathrm{FM}} + \Delta t \hspace{1mm} v_{t_n}^\theta \left(\widetilde X_{t_n}^{\mathrm{FM}}\right), \hspace{2px} n = 0, \ldots, N-1.
\end{equation}
Here, $\widetilde{X}_0 \sim \mathcal{N}(0, I_d)$, and we use $\widetilde{(\cdot)}$ to distinguish the Euler step sample from the random variables in ODE \eqref{eq:fm-ode}.
The superscript $\rm FM$ is used to distinguish the TM sampling steps, which is described next in \S \ref{subsec:tm_background}.

\subsection{Transition Matching (TM)}
\label{subsec:tm_background}

As outlined in \S \ref{subsec:cfm_background}, FM first learns a continuous-time velocity field $u_t$, which is subsequently discretized to generate a simulation trajectory $\{X_{t_n}\}_{n \in [N-1]}$ using \eqref{eq:fm-euler}.
In contrast, TM \citep{shaul2025:transition} generalizes this idea to the discrete-time setting by directly learning the probability transition kernel $p \left(X_{t_{n+1}} \vert X_{t_n}\right)$.
This approach enables the use of more expressive and non-deterministic kernels compared to FM. 

To model the transition kernel \( p \left(X_{t_{n+1}} \vert X_{t_n} \right) \), TM introduces an auxiliary latent variable \( V \).  
From the law of total probability, the transition kernel is then,
\begin{equation}
    \label{eq:tm-kernel}
    p \left(X_{t_{n+1}} \vert X_{t_n}\right) = \hspace{-1mm} \int \hspace{-0.5mm} p  \hspace{-0.5mm} \left(X_{t_{n+1}} \vert X_{t_n}, V\right) p \left(V \vert X_{t_n}\right) \mathrm{d}V,
\end{equation}
where $p \left(X_{t_{n+1}} \vert X_{t_n}, V\right)$ is chosen to be a deterministic function of $(X_{t_n}, V)$, and the conditional latent distribution \( p(V \vert X_{t_n}) \) is learned during training. 

To explicitly specify \( p \left(X_{t_{n+1}} \vert X_{t_n}, V \right) \), define $V \triangleq X_1 - X_0$ (henceforth referred to as the {\it difference latent}), and adopt the linear interpolation (CondOT) path, i.e., $X_{t_n} = (1 - n\Delta t)  X_0 + n\Delta t  X_1$.
Rearranging this, the transition becomes deterministic (given $V$) as,
\begin{equation}
    \label{eq:condOT_path_TM}
    X_{t_{n+1}} = X_{t_n} + \Delta t  V. 
\end{equation}
In other words, TM learns the conditional distribution $p_\theta(V \vert X_{t_n})$ using a model parametrized by $\theta$, and sampling $V \sim p_\theta(\cdot \vert X_{t_n})$ drives each discrete transition.

During training, the posterior distribution $p(V \vert X_{t_n})$ is learned using CFM loss \eqref{eq:cfm-loss}. 
Let $V_1 \triangleq X_1 - X_0$, sample $V_0 \sim \mathcal{N}(0, I_d)$, and choose a linear trajectory, $V_s = (1-s)V_0 + sV_1$, where $s \in [0,1]$ is used for the continuous time index to avoid conflict with $t$ in the FM ODE formulation \eqref{eq:fm-ode}. 
Then, the TM loss is,
\begin{align}
    \label{eq:tm-loss}
    \mathcal{L}_{\mathrm{TM}}(\theta) &= \mathbb{E}_{\square}
    \big\| u_s^\theta(V_s \vert X_{t_n}) - (V_1 - V_0) \big\|^2 \nonumber \\
    &\stackrel{\rm (i)}{=} \mathbb{E}_{\square}
    \big\| u_s^\theta(V_s | Z_{t_n}) - (V_1 - V_0) \big\|^2,
\end{align}
where, $\square \equiv X_0, V_0 \sim \mathcal{N}(0, I_d)$, $X_1 \sim p_1$, $n \sim \mathcal{U}[N-1]$, $s \sim \mathcal{U}[0,1]$, and the neural network $u_s^{\theta}$ predicts the velocity field associated with $p(V \vert X_{t_n})$.
Furthermore, with a slight abuse of notation
\footnote{We reuse the notation \( u_s^\theta\) to denote an output sample of the difference latent $V$ in two contexts: When the input is $Z_{t_n}$ as in $u_s^\theta(\cdot \vert Z_{t_n})$, it refers to the flow-head, whereas $u_s^\theta(\cdot \vert X_{t_n})$ refers to flow-head plus the backbone.}, 
$\rm (i)$ follows from the design of $u_s^{\theta}(\cdot \vert \cdot)$, which is composed of a large backbone encoder $f_t^\theta$, that extracts features $Z_{t_n} \triangleq f_t^\theta(X_{t_n})$, followed by a lightweight flow head $u_s^\theta (\cdot \vert Z_{t_n})$.

Once $u_s^{\theta}$ is trained, for each \(t_n\), a sample from \(p(V \mid X_{t_n})\) is obtained by simulating the ODE using \(u_s^\theta(\cdot \mid Z_{t_n})\), where \(Z_{t_n} = f_t^\theta(X_{t_n})\) remains fixed. 
Let $S$ be the total number of inner ODE steps, $s \in [S]$ denote the inner step index, and for every $n$, consider the discretization of $[0,1]$ as $t_{n,s} = s\Delta s$ ,where $\Delta s \triangleq 1/S$.
Moreoever, let $\{\widetilde X_{t_n}^{\mathrm{TM}}\}_{n \in [N]}$ denote the samples obtained by simulating the TM dynamics according to \eqref{eq:condOT_path_TM}, and for every $n$, let $\{\widetilde{V}_{n,s}\}_{s \in [S]}$ be the samples from inner Euler steps.
Then, for $n = 0, \ldots, N-1$, and $s = 0, \ldots, S-1$, the TM samples are given by
\begin{align}
     \label{eq:tm-euler}
     &\widetilde X_{t_{n+1}}^{\mathrm{TM}} = \widetilde X_{t_n}^{\mathrm{TM}} + \Delta t \hspace{1mm} \widetilde V_{t_n}, \quad \text{where } \widetilde V_{t_n} \triangleq \widetilde V_{n,1}, \nonumber \\
     &\widetilde V_{n, s+1} = \widetilde V_{n,s} + \Delta s \hspace{1mm} u_s^{\theta}(\widetilde{V}_{n,s} \vert Z_{t_n} \hspace{-1mm}=\hspace{-1mm} f_t^{\theta}(\widetilde X_{t_n}^{\mathrm{TM}})).
\end{align}

\textbf{Connection to FM.} Thm. 1 of \citeauthor{shaul2025:transition} shows that, as the total number of steps $N \to \infty$, the TM update indeed converges to the FM update for every $n$.
In other words, the update step for TM is approximately,
\begin{equation}
    \label{eq:conn_tm_fm}
    \widetilde X_{t_{n+1}}^{\mathrm{TM}}
    \approx \widetilde X_{t_n}^{\mathrm{TM}} + \Delta t \hspace{1mm} \mathbb{E} \left[ X_1 - X_0 \vert \widetilde X_{t_n}^{\mathrm{TM}} \right]
    \stackrel{\rm (i)}{=} \widetilde X_{t_{n+1}}^{\mathrm{FM}},
\end{equation}    

where $\rm (i)$ follows from \eqref{eq:FM_expected_velocity_update}.
Since TM and FM coincide in the infinite-step limit, in this work, our focus is instead on the \emph{low-step sampling regime} in the discrete-time setting, where~\citep{shaul2025:transition} provided only empirical evidence that TM outperformed FM. 
In the sections that follow, we analyze the mechanisms and conditions that explain \emph{why} and \emph{when} TM can achieve superior performance in this regime.

\paragraph{Scaling Behavior Comparison.}
We now compare the scaling behavior of FM and TM. 
As discussed in \S \ref{subsec:cfm_background}, FM requires evaluating the entire backbone network to compute \(v_{t_n}^\theta(X_{t_n})\) at each outer step $t_n$. 
In contrast, TM computes backbone features once per outer step, i.e., \(Z_{t_n}=f_{t_n}^\theta(X_{t_n})\), after which it performs \(S\) inner steps with a lightweight flow head \(u_s^\theta(\cdot | Z_{t_n})\) to sample a difference latent $V$. 
At first glance, it seems that for a fixed number of outer steps \(N\), TM should incur a higher computational cost than FM, since it involves additional inner ODE simulations. 
However, the backbone computation in TM is amortized across all inner steps, i.e., once $Z_{t_n}$ is obtained, increasing $S$ only requires repeated lightweight flow head evaluations, which are far cheaper than full backbone passes.

Formally, let \(C_B\) denote the cost of a single backbone evaluation, (a forward pass either through the network \(v_t^\theta(\cdot)\) for FM, or the feature extractor \(f_t^\theta(\cdot)\) for TM).
In our experiments in \S \ref{sec:application}, the backbone size is comparable between FM and TM, so $C_B$ can be treated as the same.
Let \(C_H\) denote the cost of a single lightweight flow head evaluation \(u_s^\theta(\cdot \mid \cdot)\), with \(C_H \ll C_B\).
The per-sample computational costs are then,
\[
\mathcal C_{\mathrm{FM}} = N C_B,\qquad
\mathcal C_{\mathrm{TM}} = N C_B + N S C_H.
\]
From this, increasing one outer step in FM incurs a cost of \(C_B\), while the same budget in TM could increase the number of inner steps by
\begin{equation}
    \label{eq:scaling_cost}
    \Delta S = \frac{C_B}{N C_H} = \frac{\kappa}{N},
    \quad\text{where } \kappa := \frac{C_B}{C_H} \gg 1.  
\end{equation}
This comparison highlights that the computational advantage of TM is pronounced when $N$ is small.

\begin{figure}[t!]
    \centering
    \includegraphics[width=0.48\textwidth]{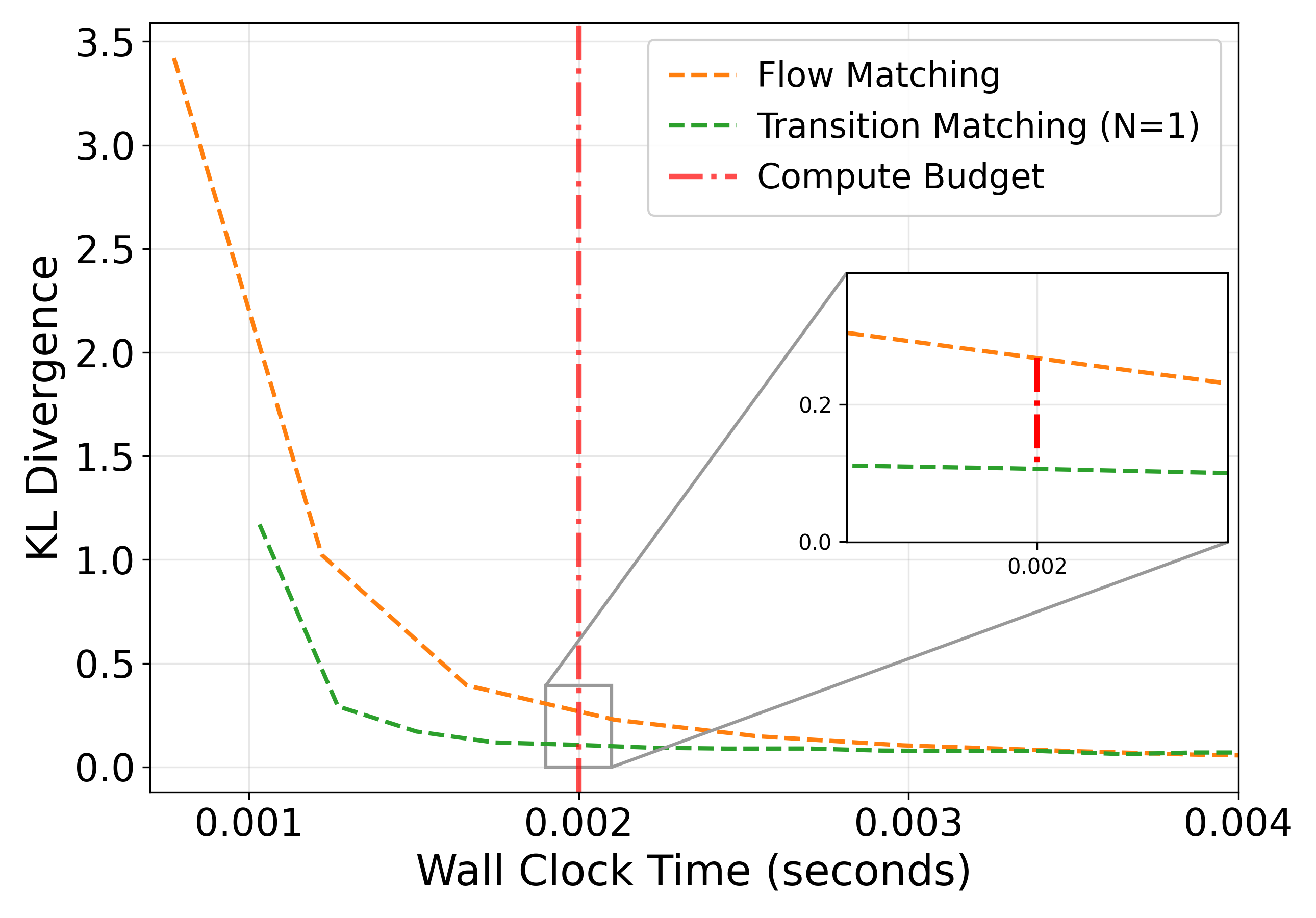}
    \caption{\textbf{KL Divergence for Unimodal Gaussian Target.}
    Comparison of scaling \(S\) at \(N=1\) for Transition Matching versus scaling \(N\) for Flow Matching. 
    TM achieves lower KL than FM under a fixed compute budget, highlighting the efficacy of increasing \(S\).
    }
    \label{fig:unimodal_kl}
\end{figure}
\begin{figure}[t]
  \centering
  \includegraphics[width=0.98\columnwidth]{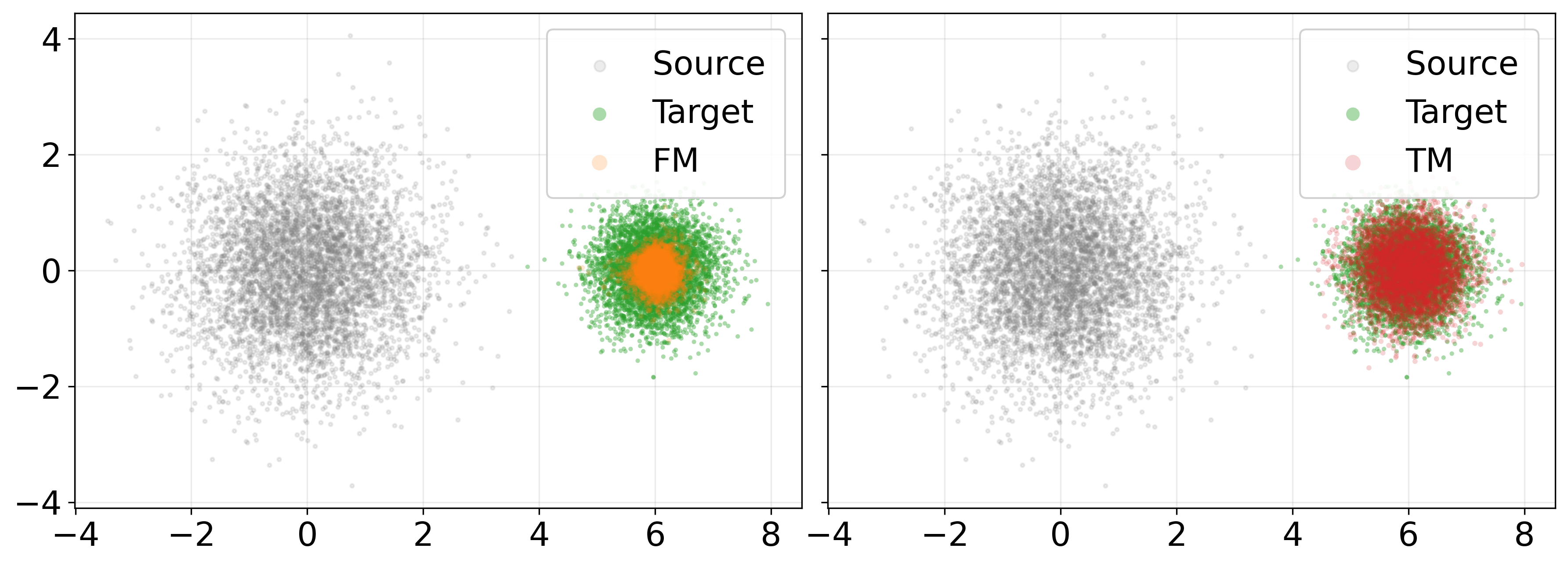}
  \vspace{-0.2cm}
  \caption{\textbf{Qualitative Visualization of Unimodal Gaussian Target.} 
  Each panel shows the source $\mathcal{N}(0, I_d)$ and target $\mathcal{N}(\mu,\sigma^2 I_d)$ distributions with the generated samples of FM (left) and TM (right). 
  With a small number of steps $(N=2)$, FM produces samples with reduced variance, 
  whereas TM $(N=1, S=2)$ preserves the target variance.}
  \label{fig:unimodal_quali}
\end{figure}

\section{UNIMODAL GAUSSIAN TARGET}
\label{sec:gauss_target_main}

As outlined in the introduction, the unimodal Gaussian target with isotropic covariance provides a simple non-trivial setting in which we explain {\it why} TM can outperform FM in the low-step sampling regime.
This setting is analytically tractable, allowing closed-form characterization of the transition kernels and explicit comparison of their updates in terms of KL divergence.
It further serves as a canonical baseline: while free of multimodal complexity, it nevertheless exposes the essential distinction between deterministic FM updates and stochastic TM updates.
Beyond offering formal guarantees, the Gaussian target case also serves as a clean testbed, demonstrating tight agreement between theoretical predictions and empirical results from toy experiments.
We consider the setting where,
\begin{equation*}
    p_0=\mathcal N(0, I_d), \qquad p_1=\mathcal N(\B{\mu}, \sigma^2 I_d).
\end{equation*}
For the linear path \eqref{eq:cond_ot}, Thm. \eqref{theorem:kl} below holds. 
\begin{theorem}
    \label{theorem:kl}
    Let \(X_0\sim\mathcal N(0,I_d)\) and \(X_1\sim\mathcal N(\mu,\sigma^2 I_d)\) with \(\sigma>0\) be independent Gaussian vectors in $\mathbb{R}^d$.
    If FM and TM iterates follow $N$ Euler steps with $S>1$ as defined in \eqref{eq:fm-euler} and \eqref{eq:tm-euler}, respectively, then we have
    \begin{equation*}
    \mathrm{KL} \left(p_1^{\mathrm{TM}}  \middle\|  \mathcal N(\mu,\sigma^2 I_d)\right)
    <
    \mathrm{KL} \left(p_1^{\mathrm{FM}}  \middle\|  \mathcal N(\mu,\sigma^2 I_d)\right),
    \end{equation*}
    where $p_1^{\rm TM}$ and $p_1^{\rm FM}$ are the marginal distributions of $\widetilde{X}_{{t_N}=1}^{\rm TM}$ and $\widetilde{X}_{{t_N}=1}^{\rm FM}$, respectively.
\end{theorem}

The proof is presented in \S \ref{subsec:gauss_target_proof}. 
The proof shows that both FM and TM exactly match the interpolation mean, so the comparison reduces to covariance. 
For any finite \(N>1\), the deterministic FM Euler update yields a covariance strictly less than the target covariance which gives a positive KL divergence. 
In contrast, TM samples a difference latent, and the induced stochasticity effectively compensates the covariance reduction. 
For any finite \(S>1\), the resulting covariance lies strictly between the FM covariance and the target covariance, so the KL divergence is smaller than that of FM.
 
\paragraph{Convergence Rate of KL Divergence.}
Understanding how quickly the KL divergence decays as the number of sampling steps $(N)$ increases provides a concrete quantitative measure of how efficiently $p_1^{\rm FM}$ or $p_1^{\rm TM}$ approach the target distribution $p_1$, thereby highlighting the tradeoff between computational cost and statistical accuracy.
Let $p_1^{\rm FM}(N)$ and $p_1^{\rm TM}(N,S)$ denote the marginal distributions of the output iterates of \eqref{eq:fm-euler} and \eqref{eq:tm-euler}, respectively, with the dependence on the outer and inner Euler steps, $N$ and $S$, made explicit. 
The following result precisely characterizes the scaling behavior of both FM and TM samplers.
\begin{corollary}
\label{rem:kl_convergence}
Within the Thm. \ref{theorem:kl}, $p_1^{\rm FM}(N)$ satisfies
\begin{equation}
    \mathrm{KL} \left(p_1^{\mathrm{FM}}(N)  \middle\|  \mathcal N(\mu,\sigma^2 I_d)\right)
    = \mathcal O \left(\frac{1}{N^2}\right),
\end{equation}
and $p_1^{\rm TM}(N,S)$ satisfies
\begin{equation}
    \mathrm{KL} \left(p_1^{\mathrm{TM}}(N,S)  \middle\|  \mathcal N(\mu,\sigma^2 I_d)\right) = \mathcal O \left(\frac{1}{N^2S^2}\right).
\end{equation}
\end{corollary}
This result follows directly from Thm.~\ref{theorem:kl}, with the proof provided in \S \ref{app:proof_kl_div_decay_rate}.
Cor.~\ref{rem:kl_convergence} underscores a key distinction: for FM, the KL divergence vanishes only as the number of outer steps \(N\) increases, whereas for TM, convergence can instead be achieved by increasing the number of inner ODE steps \(S\), even with \(N\) held fixed. 
As discussed in \S \ref{subsec:tm_background}, since scaling \(S\) in TM is generally more cost-effective than scaling \(N\) in FM, this suggests that TM can consistently achieve lower KL divergence under a fixed compute budget.

\paragraph{KL Divergence Across Discretization Steps.}
To validate the analysis, Fig.~\ref{fig:unimodal_kl} reports the KL divergence between the distributions induced by FM and TM for the target \(\mathcal{N}(\mu, \sigma^2 I_d)\), illustrating their convergence behavior. 
The horizontal axis indicates the wall clock time in seconds required to sample the target datapoints. 
For FM, we vary \(N\), whereas for TM, we fix \(N=1\) and vary only \(S\). 
Note that since \(C_B > C_H\), this setting yields a lower compute cost for TM in the low-step sampling regime (small $N$). 

The empirical curves follow our analysis in Cor.~\ref{rem:kl_convergence}. 
While both samplers converge to zero KL divergence as \(N\) and \(S\) increase, TM yields lower KL divergence than FM in the low-step sampling regime. 
Additionally, Fig.~\ref{fig:unimodal_quali} visualizes the samples generated by FM (\(N=2\)) and TM (\(N=1, S=2\)). 
We observe that FM produces noticeably smaller sample variance, whereas TM better preserves the target distribution variance even with smaller \(N\), aligning with the quantitative KL divergence trends from Thm.~\ref{theorem:kl}.

\paragraph{Analysis of Variance.}
In the unimodal Gaussian case, the distribution $p(V \mid X_{t_n})$ used to sample difference latent in \eqref{eq:tm-euler} is itself Gaussian with covariance $\operatorname{Cov}(V \mid X_{t_n}) = \frac{\sigma^2}{(1-{t_n})^2 + \sigma^2 {t_n}^2} I_d$, as noted in the proof of Thm. \ref{theorem:kl} \eqref{eq:cond-gauss}. 
When the target variance $\sigma \to 0$, the covariance of the difference latent distribution correspondingly approaches zero, and the distribution effectively behaves as a Dirac delta distribution at its mean. 
Consequently, each sampled $\widetilde{V}_{t_n}$ of TM becomes nearly identical, tightly concentrating around its expectation $\mathbb{E}[V|X_{t_n}]$. 
Following \eqref{eq:conn_tm_fm}, each TM update thus reduces to the FM update.

This delineates a precise regime in which TM achieves superior performance over FM in the low-step sampling setting for a unimodal Gaussian target with isotropic covariance $\sigma^2 I_d$ ($\sigma^2>0$): {\it stochastic sampling of the difference latent in TM preserves target variance, whereas deterministic FM updates underestimate it}. 

\begin{figure*}[t!]
    \centering
    \includegraphics[width=\linewidth]{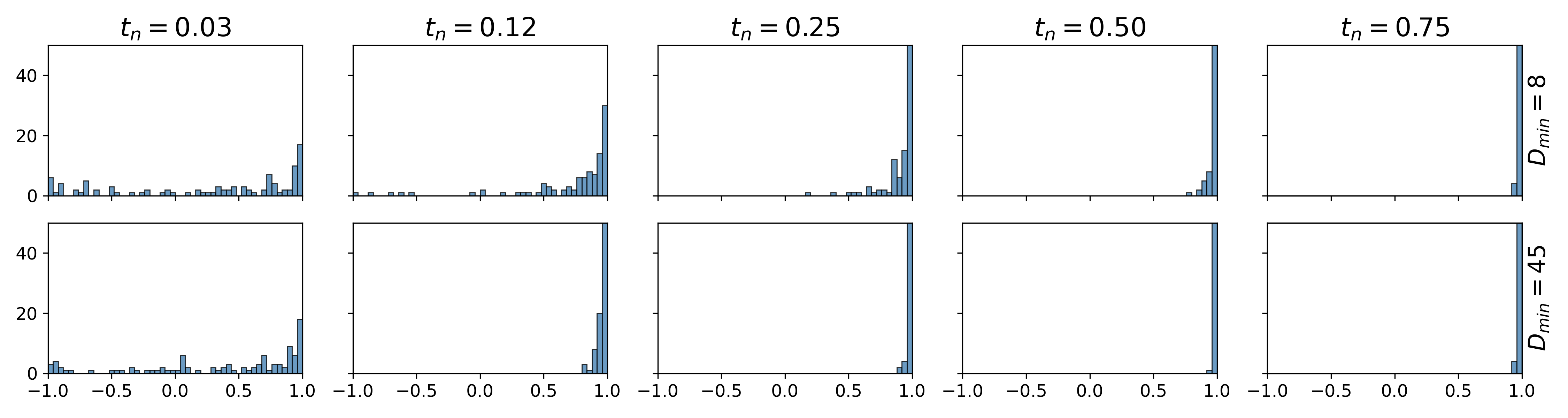}
    \caption{
        \textbf{
            Effect of $D_\text{min}$ on $p(V|X)$. 
        }
        Visualization of $p(V|X)$ using cosine-similarity histograms between difference latent samples $\widetilde V^{(m)}_{t_n} \sim p(V\mid X_{t_n})$ and $\mathbb{E}[V\mid X_{t_n}]$ for $D_{\min} \in \{8,45\}$. 
        For $D_{\min}=8$ (top) the distribution remains multimodal, whereas for $D_{\min}=45$ (bottom) it concentrates near $1$ at earlier $t_n$, indicating unimodality. 
        A larger $D_{\min}$ tightens Cor.~\ref{cor:localized-bounds}, so at a fixed $t_n$ the mixture is closer to $p(V|X, Z=k)$.
    }
    \vspace{-5pt}
    \label{fig:vel_cos_sim}
\end{figure*}


\section{MIXTURE OF GAUSSIANS TARGET}
\label{sec:multimodal_main}

Building on the unimodal Gaussian analysis in \S\ref{sec:gauss_target_main}, we now extend the framework to mixture of Gaussians and identify more precisely \emph{when} TM can yield superior performance over FM. 
Here, we consider the source distribution $p_0 = \mathcal{N}(0, I_d)$, and the target distribution,
\[
p_1 = \sum_{k=1}^K \pi_k  \mathcal N(\mu_k, \sigma_k^2 I_d),
\qquad \sum_{k=1}^K \pi_k=1, \hspace{1mm} \pi_k>0. 
\]
Unlike the unimodal Gaussian setting, the difference latent distribution \(p(V\mid X_t=x)\) is itself a mixture of Gaussian~\citep{bertrand2025:closed, Karras2022:EDM}. 

The key observation here is that along parts of the trajectory where a single component dominates (\ie~\(x\) lies near the interpolation path mean \(t\mu_k\)), the mixture distribution of difference latent $p(V|X_t=x)$ closely approximates the corresponding $k^{\rm th}$ component unimodal distribution $p(V|X_t=x, Z=k)$. 
In other words, this local reduction enables us to leverage the unimodal analysis in \S \ref{sec:gauss_target_main}. 
Formally, Prop.~\ref{prop:local-unimodal-main} quantifies the discrepancy of difference latent distribution conditioned on multimodal and unimodal targets.

\begin{proposition}
\label{prop:local-unimodal-main}
Let $X_0\sim\mathcal N(0,I_d)$ and 
\[
X_1 \sim \sum_{j = 1}^K \pi_j \mathcal{N}(\mu_j, \sigma_j^2I_d),
\quad \sum_j \pi_j=1, \hspace{1mm} \pi_j>0,
\]
with $X_0\perp X_1$. 
Under the linear interpolation path of \eqref{eq:cond_ot}, denote 
$B_t(j) = (1-t)^2 + t^2 \sigma_j^2$
to be the variance of $X_t$ conditioned on $Z = j$, i.e., $X_1$ being drawn from $j^{\rm th}$ component. 
Given $x \in \mathbb{R}^d$, let $k_t(x) \in {\rm argmin}_{j}\lVert x - t\mu_j\rVert$ (ties resolved arbitrarily), and let $D_t(x) \triangleq \lVert x - t\mu_{k_t(x)}\rVert$.
Furthermore, define the associated margin,
\begin{equation*}
    \rho_t(x) \triangleq {\rm min}_{j \neq k_t(x)} \left(\lVert x - t\mu_j \rVert - \lVert x - t\mu_{k_t(x)} \rVert\right),
\end{equation*}
to denote the gap between the distances to the closest and the second-closest path means.
Then, 
\begin{small}
    \begin{align*}
    &\lVert p(V \vert X_t = x) - p(V \vert X_t = x, Z = k_{t}(x)) \rVert_{\rm TV} \\ 
    &\leq C_{\pi}(x) \hspace{-1mm}\left(\tfrac{B^{\rm max}_t}{B^{\rm min}_t}\right)^{\hspace{-0.5mm}\frac{d}{2}}\hspace{-1mm}{\rm exp}\hspace{-0.5mm}\left(\frac{D_t^2(x)}{2}\left(\frac{1}{B^{\rm min}_t} \hspace{-0.5mm} - \hspace{-0.5mm}\frac{1}{B^{\rm max}_t}\right) \hspace{-0.5mm} - \hspace{-0.5mm}\frac{\rho_t^2(x)}{2B^{\rm max}_t}\right),  
\end{align*}
\end{small}
\hspace{-1mm}where $V = X_1 - X_0$ as before, $C_{\pi}(x) = \pi^{-1}_{k_t(x)}-1$, $B^{\rm min}_t \triangleq {\rm min}_j B_t(j)$, and $B^{\rm max}_t \triangleq {\rm max}_j B_t(j)$.
\end{proposition} 
The proof is presented in \S \ref{subsec:unimodal_approx_diff_appendix}. 
Prop.~\ref{prop:local-unimodal-main} establishes that the posterior of $V$ given $X_t=x$ can be approximated by the single component conditional $p(V \mid X_t=x, Z=k_t(x))$, which follows the unimodal Gaussian case discussed in \S\ref{sec:gauss_target_main}. 
The approximation error depends on the mixture weight of the nearest component and the variance $B_t(j)$. 
Additionally, the bound tightens when $x$ lies near its path mean (small $D_t(x)$) and the closest component is well-separated from others (large $\rho_t(x)$), in which case non-nearest components are exponentially suppressed and the posterior concentrates around a single mode. 

We next consider a special case, when the variances are comparable, formalized as $B_t^{\rm min}/B_t^{\rm max} \leq 1 - \delta$ for arbitrarily small $\delta \geq 0$, and $x$ lies within a local neighborhood of its nearest path mean, $t\mu_{k_t(x)}$.
For clarity of exposition, Cor. \ref{cor:localized-bounds} considers the case $\delta = 0$.
\begin{corollary}
\label{cor:localized-bounds}
Under the setting of Prop.~\ref{prop:local-unimodal-main}, define the minimal mean separation $D_{\min}\triangleq \min_{j\ne k}\|\mu_j-\mu_k\|$ and the neighborhood $\mathcal{E}_t \hspace{-0.5mm} \triangleq \hspace{-0.5mm} \big\{x \in \mathbb{R}^d \hspace{-0.5mm} : \hspace{-0.5mm} D_t(x) \le \sqrt{B_t^{\min}} \big\}$. Assuming $B_t^{\max} = B_t^{\min}$, for any $x \in \mathcal{E}_t$, 
\begin{align*}
    \label{cor:localized-bounds}
    \nonumber
    \big\|p(V\mid X_t=x) &- p(V\mid X_t=x, Z=k_t(x))\big\|_{\mathrm{TV}} \\
    \nonumber
    &\leq 
    C_{\pi}(x)
    \exp \left(
    2 - \frac{t^2 D_{\min}^2}{4B_t^{\max}}
    \right).
\end{align*}
\end{corollary}
The proof is provided in \S\ref{app:effect_of_mode_separation}, building on Prop.~\ref{prop:local-unimodal-main}. 
We note that the approximation error decreases with larger mean separation, $D_{\min}$, particularly as $t \to 1$. 

Additionally, smaller variances further reduce the approximation error: as $\sigma \to 0$, we have $B_t = (1-t)^2 + t^2 \sigma^2 \to 0$ for $t$ close to $1$.  
However, this regime is of little interest, since it makes FM and TM to behave essentially identically, as discussed in \S\ref{sec:gauss_target_main}.

\paragraph{Effect of Mean Separation on Sampling.}
\label{subsec:var_analysis}
To better understand multimodal targets, we study how the minimal mean separation $D_\text{min}$ in Cor.~\ref{cor:localized-bounds} governs the sampling dynamics.  
We set the target distribution \(p_1\) to a Gaussian mixture following~\citep{Cardoso2024:MCGDiff} with predefined means and identity covariances. 
To analyze the sampling dynamics, we investigate the difference latent distribution, $p(V|X_{t_n})$. 
At each timestep $t_n$, we sample the difference latent $M$ times to get $\{\widetilde V^{(m)}_{t_n}\}_{m=1}^M \sim p_\theta(V \mid X_{t_n})$ using the TM flow head in \eqref{eq:tm-euler}. 
We visualize the empirical distribution by computing the histogram of the cosine similarity between each $\widetilde V^{(m)}_{t_n}$ and the conditional expectation $\mathbb{E}[V \mid X_{t_n}]$ obtained from the samples, and plotting it across timesteps along the sampling trajectory. 

We present the result in Fig.~\ref{fig:vel_cos_sim} for two settings with $D_\text{min} \in \{8, 45 \}$. 
When $D_\text{min} = 8$ (top row), the distribution exhibits multimodal behavior until $t_{n} = 0.25$, showing weak indication of the unimodal distribution approximation. 
For larger $D_{\min}=45$ (bottom row), Cor.~\ref{cor:localized-bounds} yields a tighter bound, and thus the upper bound approximation error at a fixed timestep $t_n$ is smaller than for lower $D_{\min}$. 
Empirically, we observe the cosine similarities clustering near $1$ with non-negligible variance at earlier $t_n$. 
We present additional results analyzing the effect of target variance in \S \ref{sec:additional_results_2}. 

\paragraph{KL Divergence Comparison.}
Building on the analysis, we now compare the KL divergence of FM and TM. 
First, Lem.~\ref{lemma:high_probability_good_region} shows that once a trajectory enters the good region of a mode with sufficient margin, the dynamics can be locally approximated by the unimodal case in \S\ref{sec:gauss_target_main}. 
Specifically, if we define ${\cal G}_t(r, \rho^*) \triangleq \left\{x : \lVert x - t\mu_{k_t(x)}\rVert \leq r \text{ and } \rho_t(x) \geq \rho^* \right\}$ for some $r, \rho^*$, then $\mathbb{P}(X_t \notin {\cal G}_t(r, \rho^*))$ decreases to zero exponentially as $D_\text{min}$ increases. 
Then for $x \in {\cal G}(r, \rho^*)$, the Gaussian mixture can be closely approximated by a single unimodal component, with relative error that decays exponentially as $r$ decreases or $\rho^*$ increases. 
In this regime, the sampling dynamics approximate the unimodal Gaussian case described in \S \ref{sec:gauss_target_main}, where Thm.~\ref{theorem:kl} shows that the stochastic updates of TM better preserve the target variance and outperform FM. 
Formally, we show that the advantage of TM in the unimodal case extends to the mixture of Gaussians setting.
\begin{theorem}
    \label{thm:gaussian_mixture_KL_convergence}
    For any $M \in \{0, \ldots, N-1\}$ and $\beta \in \left(0,\frac{1}{2}\right)$, set $r = \beta t_MD_{\min}$ and $\rho^* = (1 - 2\beta)t_MD_{\min}$.
    Suppose $\widetilde{X}_{t_M}^{(\cdot)} \in \mathcal{G}_{t_M}(r, \rho^*)$, where $(\cdot)$ is either FM or TM, and $r$ and $\rho^*$.
    Then,
    \begin{align*}
        {\rm KL}(p_1^{\rm TM} \| p_1) < {\rm KL}(p_1^{\rm FM} \| p_1) - \gamma,
    \end{align*}
    where $\gamma$ can be arbitrarily close to $0$, and $p_1^{(\cdot)}$ are the marginal distributions of the final iterates.
\end{theorem}
The proof of the theorem presented in \S\ref{subsec:gaussian_mixture_KL_convergence_appx} shows that when mixture components are sufficiently well separated (i.e., when $D_\text{min}$ is large), trajectories that enter the good region of a mode remain confined within it with high probability throughout the remaining iterations.
As a consequence, the subsequent sampling dynamics effectively follow the unimodal Gaussian case analyzed in \S\ref{sec:gauss_target_main}, where TM provably achieves a smaller KL divergence than FM.
In the next section, we validate this theoretical result on Gaussian mixtures and large-scale real-world datasets, demonstrating both the sharpness of the theory and its practical significance.
\begin{figure}[t!]
    \centering
    \includegraphics[width=0.48\textwidth]{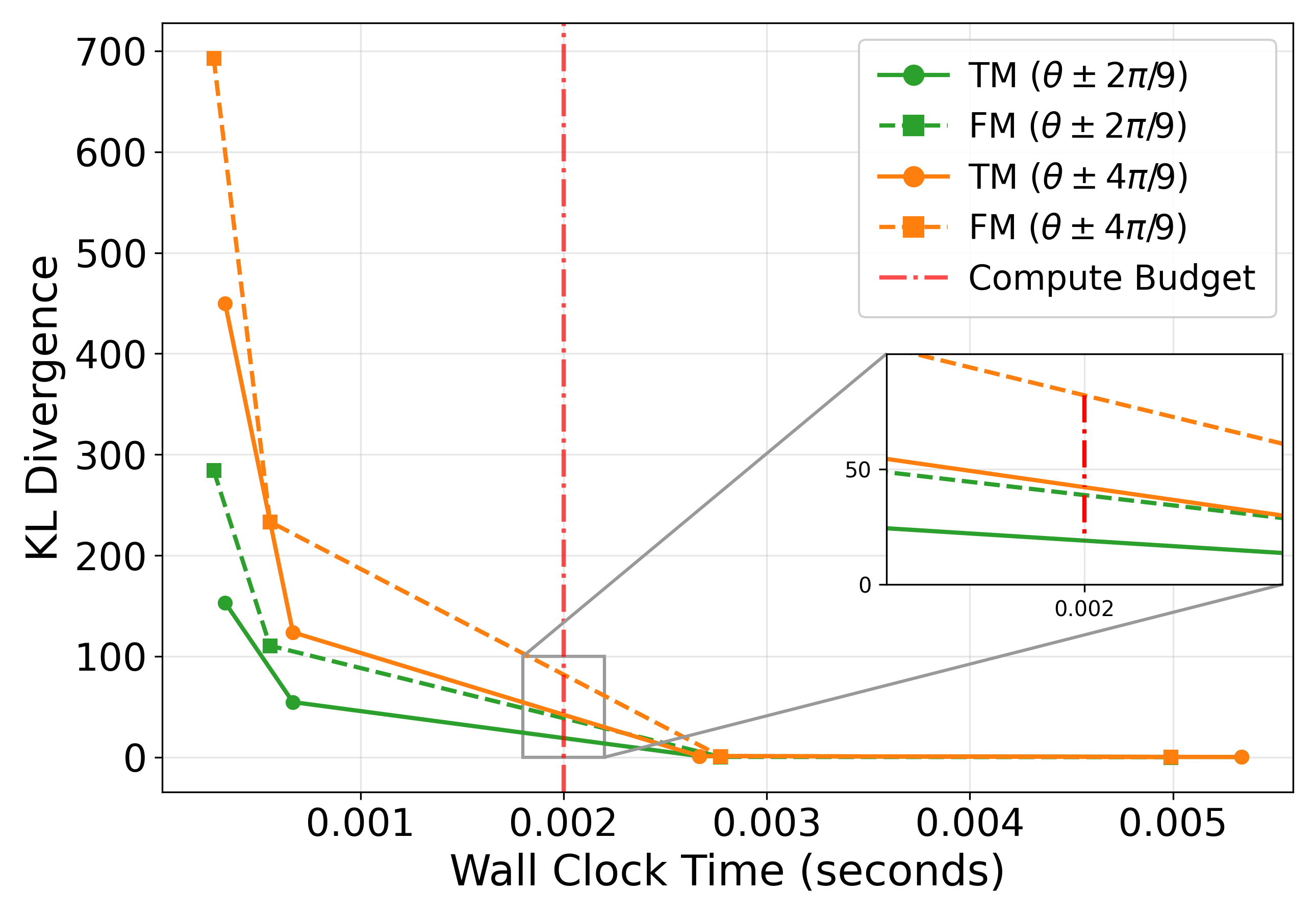}
    \caption{\textbf{KL Divergence for Mixture of Gaussians Target.} Transition Matching (TM) shows lower KL divergence than Flow Matching (FM) as the modes are more separated (orange curve, larger $D_\text{min}$). 
    The inset highlights the region near \(N = 8\). }
    \label{fig:mixture_kl}
    \vspace{-10pt}
\end{figure}
\begin{figure*}[h!]
\centering
{\small
\setlength{\tabcolsep}{0.2em} 
\def\arraystretch{0.5}

\newcolumntype{C}{>{\centering\arraybackslash}m{0.33\textwidth}} 

\begin{tabularx}{\textwidth}{C C C}
    \multicolumn{2}{c}{Class-Conditioned Image Generation} & \multicolumn{1}{c}{Frame-Conditioned Video Generation} \\


    \includegraphics[width=0.33\textwidth]{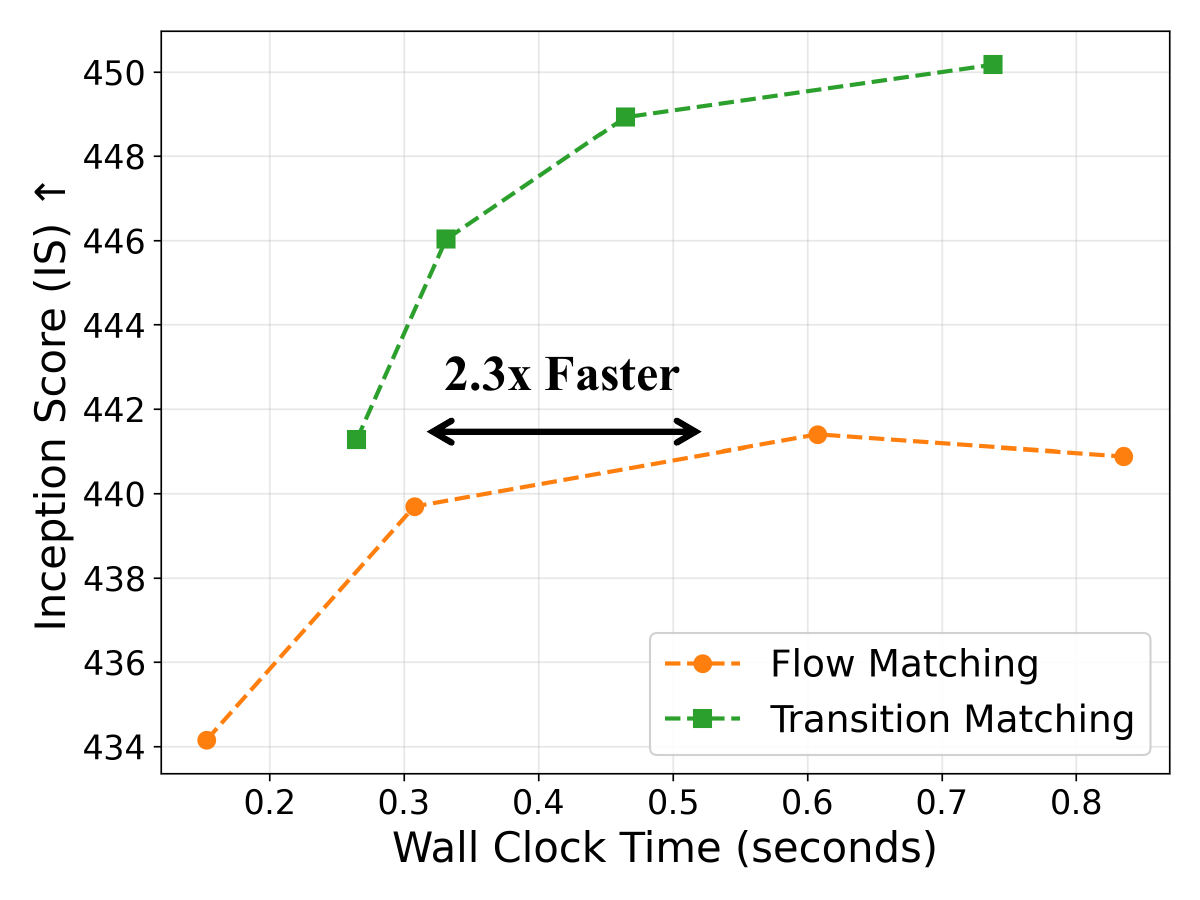} & 
    \includegraphics[width=0.33\textwidth]{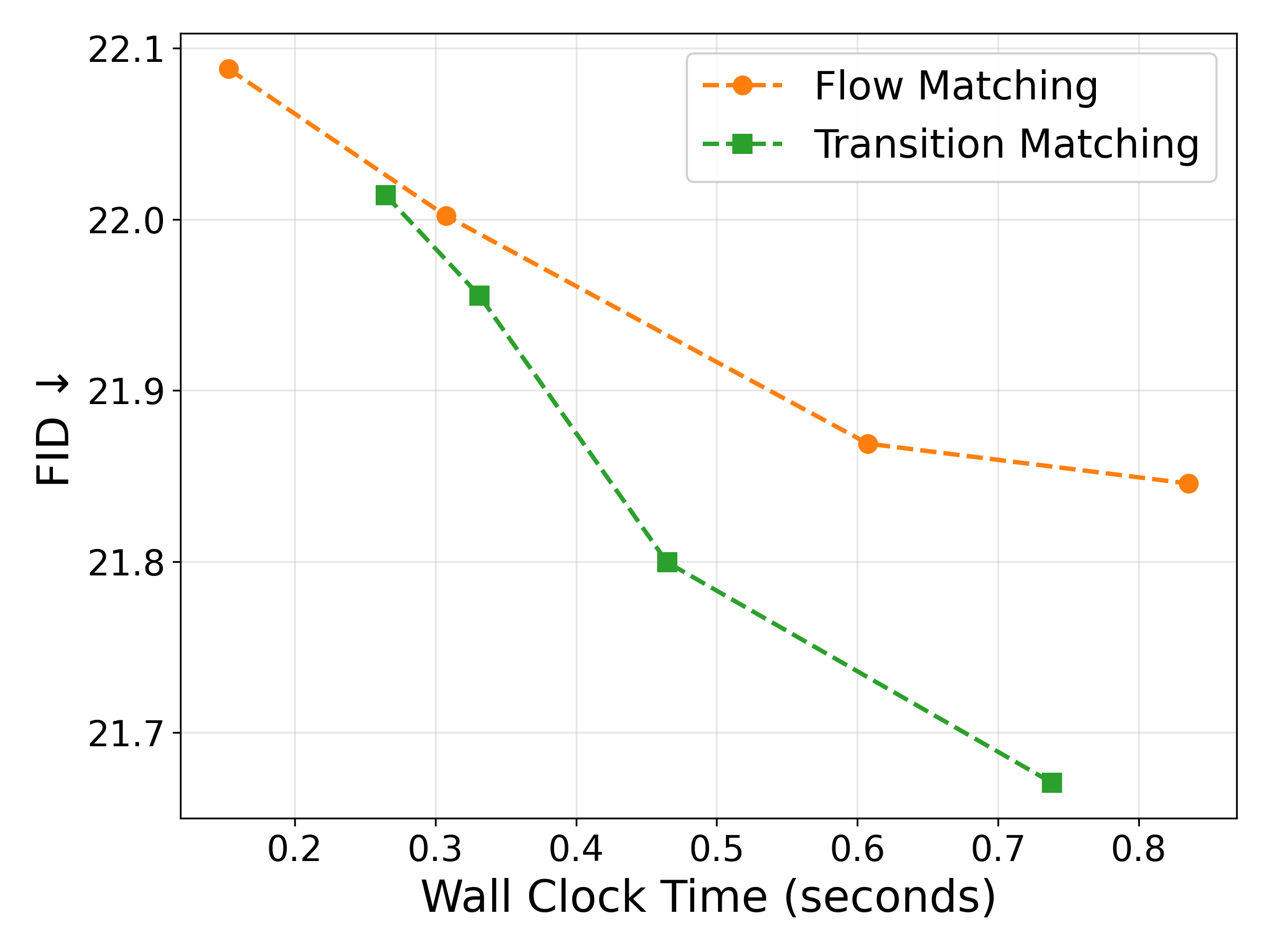} 
    & \includegraphics[width=0.33\textwidth]{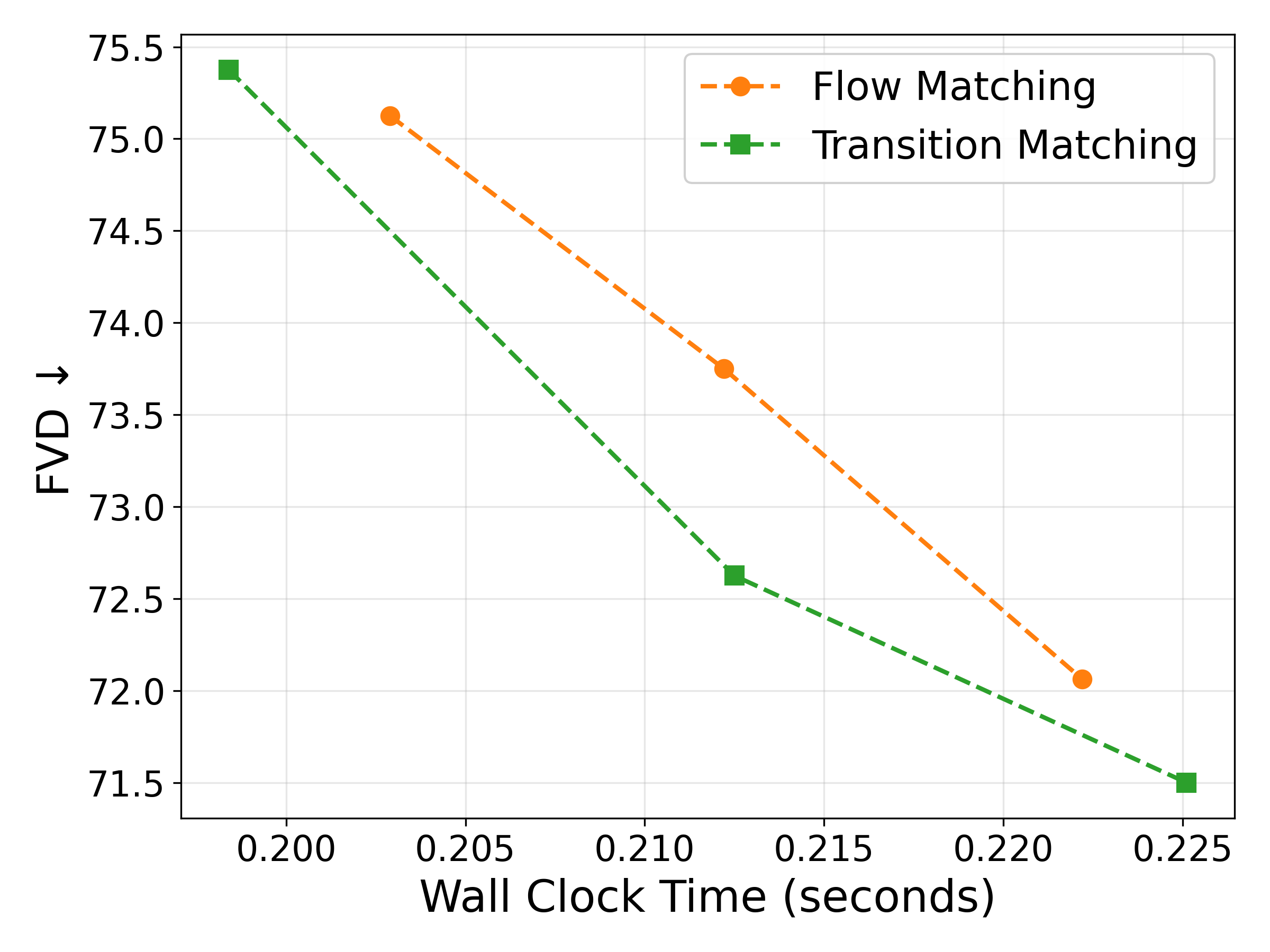} 
    
\end{tabularx}
\caption{
    \textbf{Quantitative Evaluation v.s. Wall Clock Time.} 
    Quantitative comparison of FM (orange) and TM (green) in class-conditioned image generation (left, middle) and frame-conditioned video generation (right), plotted against wall clock time measured in seconds. 
}
\label{fig:quant_pareto}
}
\end{figure*}

\section{EMPIRICAL RESULTS}
\label{sec:application}

We validate our theoretical insights on a synthetic mixture of Gaussians dataset (\S \ref{subsec:mixture_exp}), and then evaluate them on large-scale generative modeling tasks: {\it class-conditioned image generation} (\S \ref{subsec:image_gen}) and {\it frame-conditioned video generation} (\S \ref{subsec:video_gen}). 
Lastly, we present comparison results of FM and TM to diffusion-based models (\S \ref{subsec:diff_comp}). 
In Appendix, we present implementation details in \S \ref{sec:impl_details} and additional results: quantitative results with varying number of steps and qualitative results for image and video generation in \S \ref{sec:additional_results_2}.

\subsection{Experiment Setup}
\paragraph{Gaussian Synthetic Dataset.}
To validate Thm.~\ref{thm:gaussian_mixture_KL_convergence}, we consider the bimodal case (\(K=2\)) and evaluate performance under different target mean separations $D_\text{min}$ in Cor.~\ref{cor:localized-bounds}. 
We place the two component means on a circle of fixed radius at polar angles \(\theta = \pm \tfrac{2\pi}{9}\) and \(\theta = \pm \tfrac{4\pi}{9}\) with component covariances are isotropic with \( \sigma_k=0.1\). 
This results to the case where $D_{\min}=2\sin \left(\tfrac{2\pi}{9}\right)$ and $D_{\min}=2\sin \left(\tfrac{4\pi}{9}\right)$, respectively.

\paragraph{Class-Conditioned Image Generation.}
For image generation task, we compare FM~\citep{Lipman:2023CFM} and TM~\citep{shaul2025:transition} by training both on ImageNet-10K \citep{deng2009:imagenet} in a class-conditional setting at $256 \cross 256$ resolution.
The images are encoded into latents using KL-16~\citep{Rombach:2022LDM}, and training is carried out for $400$ epochs for both models. 

\paragraph{Frame-Conditioned Video Generation.}
For the video generation task, we compare TM with History-Guided Diffusion (HGD)~\citep{song2025:history}. 
HGD is a variant of FM that only differs by the interpolation path\footnote{HGD utilizes Variance Preserving~\citep{Ho:2020DDPM} interpolation path.}. 
For notational conciseness, we therefore refer to HGD as FM. 
Models are trained on Kinect-600~\citep{shaul2023kinetic} at $128 \times 128$ resolution, where videos are encoded using a pretrained VAE. 
Following \citep{song2025:history}, we train the model for $360$K steps to predict three future frames conditioned on two ground truth reference frames. 
At test time, the model conditions on a single frame and generates four future frames. 

For all experiments, we adopt DiT~\citep{peebles2023:dit} for backbone models, and the flow head of TM is implemented using MLP layers.

\subsection{Gaussian Synthetic Dataset}
\label{subsec:mixture_exp}
Fig.~\ref{fig:mixture_kl} presents the Monte Carlo estimates of the KL divergence between samples generated by FM and TM and the target mixture, plotted against wall clock time.
Consistent with Thm.~\ref{thm:gaussian_mixture_KL_convergence}, TM attains lower KL divergence than FM in the low-step regime (wall clock time $< 0.002$) for both separations $D_{\min}$ ($\theta=\pm 2\pi/9$, green; $\theta=\pm 4\pi/9$, orange).
Moreover, under fixed compute (red dotted line), the performance gap between FM and TM is greater for the larger $D_{\min}$, indicating that stronger local unimodality associated with larger $D_{\min}$ yields a structural advantage for TM via better variance preservation, and consequently, FM requires more steps to match TM.

\subsection{Class-Conditioned Image Generation}
\label{subsec:image_gen}
We evaluate FM and TM on $50,000$ images from the test set and report Inception Score (IS)~\citep{salimans2016:is} and Fréchet Inception Distance (FID)~\citep{heusel2017:fid} to assess sample quality and diversity, where FID is computed between generated samples and the ground truth test set. 
For FM, we vary the number of sampling steps $N \in \{ 16, 32, 64, 96 \}$, while for TM, we fix $N=16$, and scale $S \in \{ 2, 4, 8, 16 \}$. 
Quantitative evaluations are reported for each setting against wall clock time for generating a $256\times 256$ image. 

\paragraph{Results.}
Fig.~\ref{fig:quant_pareto} (left, middle) presents quantitative results under varying compute budgets. 
Consistent with the trend observed in \S\ref{subsec:mixture_exp}, TM outperforms FM under a fixed compute budget across both evaluation metrics, achieving a Pareto front.
Notably, TM with $S=2$ achieves comparable IS to FM with $N=64$, yielding a $2.3 \times$ wall clock time speedup (black arrow).

\subsection{Frame-Conditioned Video Generation}
\label{subsec:video_gen}
For video generation, we report Fréchet Video Distance (FVD)~\citep{unterthiner2018:fvd}, computed between generated and corresponding ground-truth videos, to assess temporal consistency and perceptual realism. 
Following the image generation experiment, we fix \( N = 16 \) for TM and scale \( S \in \{12, 16, 20\} \), while \( N \) for FM is chosen to ensure comparable compute between the two methods. 
We report quantitative results along with the wall clock time required to generate a single video. 

\paragraph{Results.}
The quantitative results are presented in Fig.~\ref{fig:quant_pareto} (right). 
Consistent with the trends observed in the mixture of Gaussians and class-conditioned image generation experiments, TM outperforms FM under a matched compute budget, achieving a more favorable quality–compute Pareto front.

\textit{Empirical Takeaway.}
From our empirical observations, TM outperforms FM under matched—or even lower—compute budgets across both large scale image and video generation tasks. 
We hypothesize that conditional inputs (\eg~class labels, reference frames) reshape the target into well separated modes with non-negligible component variance. 
In this regime, TM with stochastic difference latent updates better preserves component variance, while deterministic FM updates tend to underestimate. 
As a result, TM captures the target distribution more faithfully and achieves higher sample quality under comparable compute.

\subsection{Comparison to Diffusion Models}
\label{subsec:diff_comp}
We compare TM with three distinct sampling strategies: DDPM (Ancestral Sampling)~\cite{Ho:2020DDPM}, DDIM (Skip-Step Sampler)~\cite{Song2020:DDIM}, and Optimal Covariance Matching~\cite{ou2024:ocm}. 
We present details of each solvers used in each baseline in Appendix \S \ref{sec:appx_diff_comp}.

In Tab.~\ref{tab:comparison_diff}, we present the KL divergence of these samplers on a unimodal Gaussian target. For DDPM~\cite{Ho:2020DDPM}, DDIM~\cite{Song2020:DDIM}, OCM~\cite{ou2024:ocm}, and FM, we scale the number of integration steps $N$. 
For TM, we fix $N=1$ and scale $S$. 
Note that this regime affords TM a lower total compute budget than the baselines at each case. 
Additionally, for OCM, we utilize the ground truth optimal covariance which can be derived in closed form for this unimodal Gaussian case.

\begin{table}[h!]
    \renewcommand{\arraystretch}{1.5} 
    \setlength{\tabcolsep}{4pt}
    \scriptsize
    \centering
    \caption{\textbf{Quantitative Comparison of FM and TM to Diffusion Models.}}
    \label{tab:comparison_diff}
    \newcolumntype{Y}{>{\centering\arraybackslash}X}
    \begin{tabularx}{\linewidth}{l *{6}{Y}}
        \toprule
        \multicolumn{7}{c}{\textbf{Number of Steps}} \\
        \midrule
        Model & 2 & 4 & 8 & 16 & 128 & 1000 \\
        \midrule
        \makecell[l]{DDPM \\ \cite{Ho:2020DDPM}} & 73.81 & 72.89 & 71.00 & 68.30 & 36.38 & 0.00 \\
        \makecell[l]{DDIM \\ \cite{Song2020:DDIM}}    & 6.88 & 1.42 & 0.40 & 0.15 & 0.01 & 0.00 \\
        \makecell[l]{OCM \\ \cite{ou2024:ocm}}    & 7.48 & 0.01 & 0.00 & 0.00 & 0.00 & 0.00 \\
        \makecell[l]{FM \\ \cite{Gat:2024DFM}}    & 0.92 & 0.21 & 0.06 & 0.02 & 0.00 & 0.00 \\
        \makecell[l]{TM $(N=1)$ \\ \cite{shaul2025:transition}}   & 0.28 & 0.11 & 0.05 & 0.04 & 0.01 & 0.00 \\
        \bottomrule
    \end{tabularx}
\end{table}


As expected, classic DDPM ancestral sampling incurs significant errors when skipping steps, as its transition kernel is derived under the assumption of Markov process. 
While DDIM mitigates this by reformulating the reverse process, it still exhibits non-negligible divergence in the few-step regime ($N \leq 8$); this stems from its heuristic variance choice, which fails to capture the true conditional covariance of the strided transition. 
In contrast, OCM achieves near-perfect convergence even at very few steps ($N=4$) by leveraging the ground truth optimal covariance. 
However, it is important to note that OCM utilized the ground truth Hessian and covariance for Gaussian targets. 
In practical scenarios, the ground truth covariance is intractable, forcing OCM to rely on approximations (e.g., via the Hutchinson estimator). 
On the other hand, TM relaxes the explicit Gaussian assumption and trains a continuous flow to directly model the transition kernel. 

\section{CONCLUSION}
\label{sec:conclusion}
In this work, we provided a theoretical results and analysis that explain when and why Transition Matching (TM) outperforms Flow Matching (FM). 
In unimodal Gaussian targets, we proved that TM achieves strictly lower KL divergence than FM and converges faster under fixed compute. 
We extend the analysis to Gaussian mixtures, identifying locally unimodal regimes in which the sampling dynamics approximate the unimodal case, where TM outperforms FM. 
Together, these results establish conditions under which TM should be preferred: targets with well-separated modes and non-negligible variances. Empirical studies on controlled Gaussian settings confirm the theoretical predictions, while experiments on real-world image and video generation further validated the practical impact.

\section*{Acknowledgements}
We thank Kyeongmin Yeo for providing constructive feedback on the analysis of the unimodal Gaussian setup. 
We are also grateful to Tao Yu and Kaan Özkara for maintaining the server clusters and offering valuable advice on neural network optimization behaviors. 
This work was supported by the National Research Foundation of Korea (NRF) (RS-2026-25486000); the Institute of Information \& Communications Technology Planning \& Evaluation (IITP) grants (RS-2024-00399817, RS-2025-25441313, RS-2025-25443318), funded by the Korean government (MSIT); the Industrial Technology Innovation Program (RS-2025-02317326), funded by the Korean government (MOTIE); the National Supercomputing Center (KSC-2025-CRE-0475); and the DRB-KAIST SketchTheFuture Research Center. 

{
    \small
    \bibliographystyle{plainnat}
    \bibliography{main}
}

\section*{Checklist}



\begin{enumerate}

  \item For all models and algorithms presented, check if you include:
  \begin{enumerate}
    \item A clear description of the mathematical setting, assumptions, algorithm, and/or model. \textcolor{blue}{[Yes, we present assumptions and setting for all our proofs \S \ref{subsec:gauss_target_proof}-\ref{subsec:gaussian_mixture_KL_convergence_appx}]}
    \item An analysis of the properties and complexity (time, space, sample size) of any algorithm. \textcolor{blue}{[Yes, we present time complexity \S \ref{sec:gauss_target_main}]}
    \item (Optional) Anonymized source code, with specification of all dependencies, including external libraries. \textcolor{blue}{[No. We will open-source our code later.]}
  \end{enumerate}

  \item For any theoretical claim, check if you include:
  \begin{enumerate}
    \item Statements of the full set of assumptions of all theoretical results. \textcolor{blue}{[Yes, assumptions and proofs are presented with assumptions \S \ref{subsec:gauss_target_proof}-\ref{subsec:gaussian_mixture_KL_convergence_appx}]}
    \item Complete proofs of all theoretical results. \textcolor{blue}{[Yes]}
    \item Clear explanations of any assumptions. \textcolor{blue}{[Yes]}
  \end{enumerate}

  \item For all figures and tables that present empirical results, check if you include:
  \begin{enumerate}
    \item The code, data, and instructions needed to reproduce the main experimental results (either in the supplemental material or as a URL). \textcolor{blue}{[Yes, we present instructions to reproduce the results. We also commit to open-sourcing our code later.]}
    \item All the training details (e.g., data splits, hyperparameters, how they were chosen). \textcolor{blue}{[Yes, training configurations are presented]}
    \item A clear definition of the specific measure or statistics and error bars (e.g., with respect to the random seed after running experiments multiple times). \textcolor{blue}{[Yes. It is presented in \S \ref{sec:additional_results_2}.]}
    \item A description of the computing infrastructure used. (e.g., type of GPUs, internal cluster, or cloud provider). \textcolor{blue}{[Yes, compute resources are specified.]}
  \end{enumerate}

  \item If you are using existing assets (e.g., code, data, models) or curating/releasing new assets, check if you include:
  \begin{enumerate}
    \item Citations of the creator If your work uses existing assets. \textcolor{blue}{[Yes]}
    \item The license information of the assets, if applicable. \textcolor{blue}{ImageNet is available for non-commercial research and educational use, while the Kinetics dataset is distributed under the permissive Creative Commons Attribution 4.0 (CC BY 4.0) license.}
    \item New assets either in the supplemental material or as a URL, if applicable. \textcolor{blue}{[Not Applicable]}
    \item Information about consent from data providers/curators. \textcolor{blue}{[Not Applicable]}
    \item Discussion of sensible content if applicable, e.g., personally identifiable information or offensive content. \textcolor{blue}{[Not Applicable]}
  \end{enumerate}

  \item If you used crowdsourcing or conducted research with human subjects, check if you include:
  \begin{enumerate}
    \item The full text of instructions given to participants and screenshots. \textcolor{blue}{[Not Applicable]}
    \item Descriptions of potential participant risks, with links to Institutional Review Board (IRB) approvals if applicable. \textcolor{blue}{[Not Applicable]}
    \item The estimated hourly wage paid to participants and the total amount spent on participant compensation. \textcolor{blue}{[Not Applicable]}
  \end{enumerate}

\end{enumerate}

\ifpaper
\else
    \clearpage
    \newpage
    \onecolumn
    \appendix
    \section*{APPENDIX}

\def\thesection{\Alph{section}}

\section{PROOFS OF MAIN RESULTS}
\label{app:proofs}
\subsection{Proof of Thm. \ref{theorem:kl}: Gaussian Target Distribution}
\label{subsec:gauss_target_proof}

\begin{theorem}
\label{thm:tm_le_cfm_kl}
Let \(X_0\sim\mathcal N(0,I_d)\) and \(X_1\sim\mathcal N(\mu,\sigma^2 I_d)\) with \(\sigma>0\) be independent Gaussian vectors in $\mathbb{R}^d$.
Consider the discretization of $[0, 1]$ as \(t_n=n\Delta t\), \(\Delta t=1/N\) with \(N>1\), and let $S > 1$ be the number of inner ODE steps for TM in \eqref{eq:tm-euler}. 
If FM and TM iterates are updated according to \eqref{eq:fm-euler} and \eqref{eq:tm-euler}, respectively, then
\[
\mathrm{KL} \left(p_1^{\mathrm{TM}}  \middle\|  \mathcal N(\mu,\sigma^2 I_d)\right)
<
\mathrm{KL} \left(p_1^{\mathrm{FM}}  \middle\|  \mathcal N(\mu,\sigma^2 I_d)\right),
\]
where $p_1^{\rm TM}$ and $p_1^{\rm FM}$ are the marginal distributions of $\widetilde{X}_{t_N=1}^{\rm TM}$ and $\widetilde{X}_{t_N=1}^{\rm FM}$, respectively.
\end{theorem}

\begin{proof}

For any $t \in [0,1]$, let the corresponding point on the linear trajectory from $X_0$ to $X_1$ be $X_t=(1-t)X_0+tX_1$, and let $V = X_1 - X_0$ be the difference latent .
Since $X_0 \perp X_1$, and $X_t$ is a convex combination of two independent Gaussian random variables, from the linearity of covariance, we have,
\begin{align}
    \label{eq:Cov_Xt}
    &\operatorname{Cov}(X_t) \triangleq \mathbb{E}\left[(X_t - \mathbb{E}[X_t])(X_t - {\mathbb E}[X_t])^\top\right] = B(t)I_d, \quad \text{where } B(t) \triangleq (1-t)^2+\sigma^2 t^2, \\
    \text{and,} \quad &\operatorname{Cov}(X_t,V) \triangleq \mathbb{E}\left[(X_t - \mathbb{E}[X_t])(V - {\mathbb E}[V])^\top\right] = A(t)I_d, \quad \text{where } A(t) \triangleq t(1+\sigma^2)-1.
\end{align}

Using some algebra, it can be shown that
\begin{equation}
    \label{eq:gauss-id}
    A(t)^2+\sigma^2=(1+\sigma^2)B(t).
\end{equation}

Since $X_0 \perp X_1$, the stacked random variable $(X_0, X_1) \in \mathbb{R}^{2d}$ is jointly Gaussian.
Consequently, $(X_t, V)$ is also jointly Gaussian, since it is a linear transformation of $(X_0, X_1)$. 
Using Lemma \ref{lem:joint-gaussian-conditional-distribution} and \eqref{eq:gauss-id} above, $V \vert X_t = x$ is also Gaussian-distributed, and we have, 
\begin{equation}
    \label{eq:cond-gauss}
    \mathbb E[V\vert X_t=x]=\mu+k(t)(x-\mu t),
    \hspace{2ex}
    \operatorname{Cov}(V\vert X_t=x)=\tau^2(t) I_d,
    \hspace{2ex} \text{where }
    k(t) \triangleq \frac{A(t)}{B(t)} \hspace{1ex}\text{ and }\hspace{1ex} \tau^2(t) \triangleq\frac{\sigma^2}{B(t)}.
\end{equation}

\textbf{KL Divergence for FM steps.} From the expression in \eqref{eq:fm-euler} and the fact that perfect minimization of \eqref{eq:cfm-loss} will ensure $v_{t_n}^{\theta} = \mathbb{E}[V \vert \widetilde{X}_{t_n}^{\rm FM}]$ \eqref{eq:FM_expected_velocity_update}, the deterministic update of the FM Euler step can be written recursively as
\begin{equation}
\label{eq:cfm-update}
    \widetilde X_{t_{n+1}}^{\mathrm{FM}}=a_n \widetilde X_{t_n}^{\mathrm{FM}}+b_n, \quad \text{where, } a_n \triangleq 1+\Delta t k(t_n) \text{ and } b_n \triangleq \Delta t\bigl(\mu-k(t_n)\mu t_n\bigr)
\end{equation}

Let $m_n=\mathbb E[\widetilde X_{t_n}^{\mathrm{FM}}]$. 
Then,
\[
m_{n+1}=a_n m_n+b_n,\qquad m_0=0.
\]
By induction, $m_n=\mu t_n$ for all $n$ since
\begin{equation}
    \label{eq:mean-track}
    m_{n+1}=(1+\Delta t k(t_n))\mu t_n+\Delta t(\mu-k(t_n)\mu t_n)
=\mu t_n+\Delta t \mu=\mu t_{n+1}.
\end{equation}
Therefore, the mean follows the interpolation path and leads to $m_{N} = \mu$.

Furthermore, from \eqref{eq:cfm-update}, the covariance evolves as
\begin{equation}
    \operatorname{Cov}\left(\widetilde X_{t_{n+1}}^{\mathrm{FM}}\right) = a_n^2\operatorname{Cov}\left(\widetilde X_{t_n}^{\mathrm{FM}}\right).
\end{equation}
In other words, let \(\operatorname{Cov}\left(\widetilde X_{t_n}^{\mathrm{FM}}\right)=s_n^{\mathrm{FM}} I_d\), and the evolution of $s_N^{\rm FM}$ is given by,
\begin{equation}
    \label{eq:cfm-var-rec}
    s_{n+1}^{\mathrm{FM}}=a_n^2 s_n^{\mathrm{FM}},
    \qquad
    s_0^{\mathrm{FM}}=1.
    \quad\Rightarrow\quad
    s_N^{\mathrm{FM}}=\prod_{n=0}^{N-1}\bigl(1+\Delta t k(t_n)\bigr)^2. 
\end{equation}

On the other hand, the true variance, i.e., $B(t_n)$ defined in \eqref{eq:Cov_Xt} can be written as a recursion as,
\begin{align}
\label{eq:B-update}
B(t_{n+1}) &= B(t_n)+2\Delta t A(t_n)+(\Delta t)^2(1+\sigma^2) \nonumber \\
&\stackrel{\rm (i)}{=} B(t_n) + 2\Delta t k(t_n) B(t_n) + (\Delta t)^2 \left(k(t_n)^2B(t_n) + \frac{\sigma^2}{B(t_n)}\right) \nonumber \\
&= (1+\Delta t k(t_n))^2 B(t_n)+(\Delta t)^2\frac{\sigma^2}{B(t_n)},
\end{align}
where $\rm (i)$ makes use of the notation of $k(t_n)$ along with \eqref{eq:gauss-id}. 
Define the ratio of the variance of the sample obtained from FM Euler step to the true variance of the interpolant as \(r_n^{\mathrm{FM}} \triangleq s_n^{\mathrm{FM}}/B(t_n)\). 
From \eqref{eq:cfm-var-rec} and \eqref{eq:B-update},
\begin{equation}
    \label{eq:rcfm-rec}
    r_{n+1}^{\mathrm{FM}}=w_n  r_n^{\mathrm{FM}},\qquad r_0^{\mathrm{FM}}=1,
\end{equation}
where
\begin{equation}
\label{eq:wndef}
    w_n \triangleq \frac{(1+\Delta t k(t_n))^2}{(1+\Delta t k(t_n))^2+(\Delta t)^2 \sigma^2/B(t_n)^2}\in(0,1).
\end{equation}

Hence, the {\it variance contraction factor}, \(r_n^{\mathrm{FM}}\) strictly decreases and is always less than $1$, which means,
\begin{equation}
    \label{eq:cfm-var-bound}
    s_N^{\mathrm{FM}}<B(1)=\sigma^2.
\end{equation}
With matching means \eqref{eq:mean-track} and isotropic covariances, the KL divergence \eqref{lem:KL-divergence-Gaussian} between the generated samples at $t_N = 1$ and the target is given by
\begin{equation}
    \label{eq:cfm-kl}
    \mathrm{KL} \left(p_{t_N}^{\mathrm{FM}}  \middle\|  \mathcal N(\mu,\sigma^2 I_d)\right)
    =\frac{d}{2} \left(\frac{s_N^{\mathrm{FM}}}{\sigma^2}-1-\log\frac{s_N^{\mathrm{FM}}}{\sigma^2}\right)>0.
\end{equation}
In Cor. \ref{rem:kl_convergence} (proved in \S \ref{app:proof_kl_div_decay_rate}), we show that this KL divergence converges to $0$ as $N \to \infty$ in ${\mathcal O}\left(\frac{1}{N^2}\right)$.

\textbf{KL divergence for TM steps.}
\label{sec:tm-kl}
As seen from \eqref{eq:tm-euler}, given $X_0 \sim \mathcal{N}(0, I_d)$, for every outer iteration $n \in [N-1]$, TM draws a sample from the conditional distribution of the \textit{difference latent}, $p(V \vert \widetilde{X}_{t_n}^{\rm TM})$, by iterating over $S$ inner Euler steps.
Analogous to \eqref{eq:cfm-update}, the TM updates can also be written as
\begin{equation}
    \label{eq:tm-update}
    \widetilde X_{t_{n+1}}^{\mathrm{TM}}=a_n \widetilde X_{t_n}^{\mathrm{TM}}+b_n+\eta_n,
    \qquad
    \eta_n=\Delta t \varepsilon_n, \quad
    \varepsilon_n\sim\mathcal N \bigl(0, c_S \tau^2(t_n) I_d\bigr),
\end{equation}
where the additional variance term $\eta_n$ arises from the stochasticity of the updates as a result of sampling from $p(V \vert \widetilde{X}_{t_n}^{\rm TM})$, instead of using $\mathbb{E}[V \vert \widetilde{X}_{t_n}^{\rm TM}]$ as the update direction.
If the sampling of $V \vert \widetilde{X}_{t_n}^{\rm TM}$ had been perfect, then $\operatorname{Cov}(\eta_n) = \tau^2(t_n)I_d$ from \eqref{eq:cond-gauss}.
For instance, this is the case when the number of inner iterations $S \to \infty$.
However, when a \textit{finite} number of inner Euler steps is used to simulate the inner ODE, and consequently, approximately sample $V \vert \widetilde{X}_{t_n}^{\rm TM}$, it introduces an additional scaling factor akin to \eqref{eq:rcfm-rec} -- the effect of which is captured using $c_S$.
This mirrors the FM case, as the target distribution, i.e., \eqref{eq:cond-gauss}, is also unimodal Gaussian.
In other words,
\begin{equation}
    \label{eq:finiteS-attn}
    \operatorname{Cov} \left(\widetilde V_{t_n}\mid \widetilde X_{t_n}^{\mathrm{TM}}\right)
    =c_S \tau^2(t_n) I_d,
    \qquad 0 < c_S\le 1,
\end{equation}
where $\widetilde V_{t_n}$ is defined as in \eqref{eq:tm-euler}.
Note that \(c_S \to 1\) as \(S \to \infty\) as shown in Cor.~\ref{app:proof_kl_div_decay_rate}, and the inner ODE recovers the true $p(V | \widetilde{X}_{t_n}^{\rm TM})$. 

Similar to the FM case in Equation.~\ref{eq:mean-track}, the mean follows the interpolation path. 
That is, overloading the notation $m_n$, and denoting $m_n=\mathbb E[\tilde X_{t_n}^{\mathrm{TM}}]$, we see from induction,
\[
m_{n+1}=(1+\Delta t k(t_n))\mu t_n+\Delta t(\mu-k(t_n)\mu t_n)
=\mu t_{n+1}.
\]

Let us denote \(\operatorname{Cov}(\widetilde X_{t_n}^{\mathrm{TM}})=s_n^{\mathrm{TM}} I_d\). 
From \eqref{eq:tm-update},
\begin{equation}
    \label{eq:tm-var-rec}
    s_{n+1}^{\mathrm{TM}}=a_n^2 s_n^{\mathrm{TM}}+(\Delta t)^2 c_S \tau^2(t_n)
    =a_n^2 s_n^{\mathrm{TM}}+(\Delta t)^2 c_S \frac{\sigma^2}{B(t_n)}.
\end{equation}

With \(r_n^{\mathrm{TM}} \triangleq s_n^{\mathrm{TM}}/B(t_n)\) and using \eqref{eq:B-update} and \eqref{eq:tm-var-rec},
\begin{equation}
    \label{eq:rtm-rec}
    r_{n+1}^{\mathrm{TM}}
    =\frac{(1+\Delta t k(t_n))^2 r_n^{\mathrm{TM}}+(\Delta t)^2 c_S \sigma^2/B(t_n)^2}
    {(1+\Delta t k(t_n))^2+(\Delta t)^2 \sigma^2/B(t_n)^2}
    = w_n r_n^{\mathrm{TM}}+(1-w_n)c_S,
    \qquad r_0^{\mathrm{TM}}=1,
\end{equation}
where $w_n$ is as defined in \eqref{eq:wndef}.
Combining \eqref{eq:rcfm-rec} and \eqref{eq:rtm-rec}, we get,
\begin{equation}
    \label{eq:sandwich}
    r_n^{\mathrm{FM}} < r_n^{\mathrm{TM}}\le 1
    \quad\Longleftrightarrow\quad
    s_n^{\mathrm{FM}} < s_n^{\mathrm{TM}}\le B(t_n)\quad\text{for all }n.
\end{equation}

Since the means are the same for $t_N=1$, so the KL divergence between the target distribution and the distribution of $\widetilde{X}_{t_N}^{\rm TM}$ is given by
\begin{equation}
    \label{eq:tm_kl}
    \mathrm{KL} \left(p_{t_N}^{\mathrm{TM}} \| \mathcal N(\mu,\sigma^2 I_d)\right)
    =\frac{d}{2}\left(\frac{s_N^{\mathrm{TM}}}{\sigma^2}-1-\log\frac{s_N^{\mathrm{TM}}}{\sigma^2}\right).
\end{equation}

Since \(\phi(x)=\tfrac{d}{2}(x-1-\log x)\) is strictly decreasing on \((0,1]\), \eqref{eq:cfm-kl} and \eqref{eq:tm_kl} together with the fact $t_N = 1$, imply,
\begin{equation}
    \label{eq:tm-kl-order}
    \mathrm{KL} \left(p_1^{\mathrm{TM}}  \middle\|  \mathcal N(\mu,\sigma^2 I_d)\right)
    < \mathrm{KL} \left(p_1^{\mathrm{FM}}  \middle\|  \mathcal N(\mu,\sigma^2 I_d)\right).
\end{equation}
This concludes the proof.
\end{proof}

\subsection{Proof of Corollary \ref{rem:kl_convergence}: KL divergence decay rate for FM and TM}
\label{app:proof_kl_div_decay_rate}

\begin{corollary*}
{\bf [KL divergence decay rate for FM and TM]}
Under the setting of Thm. \ref{theorem:kl} with the discretization of $[0,1]$ given by \(t_n = n\Delta t\), \(\Delta t = 1/N\), the covariance of the FM iterates, given by $s_N^{\mathrm{FM}} = \prod_{n=0}^{N-1}\left(1 + \Delta t  k(t_n)\right)^2$,
satisfies $s_N^{\mathrm{FM}} = \sigma^2 + \mathcal{O}\left(N^{-1}\right)$ in the asymptotic limit of $N \to \infty$.
Consequently, we have,
\[
\mathrm{KL}\left(p_1^{\mathrm{FM}}(N) \middle\| \mathcal{N}(\mu, \sigma^2 I_d)\right)
= \mathcal{O}\left(\frac{1}{N^2}\right).
\]
Moreover, the TM iterates obtained according to \eqref{eq:tm-euler} satisfy 
\[
\mathrm{KL} \left(p_1^{\mathrm{TM}}(N, S)  \middle\|  \mathcal N(\mu,\sigma^2 I_d)\right)
= \mathcal O \left(\frac{1}{N^2 S^2}\right)\to 0,
\]
%
%
\end{corollary*}

\begin{proof}

{\bf Asymptotic decay rate for FM}.
We begin by taking the logarithm of \(s_N^{\mathrm{FM}}\), i.e.,
\[
\log s_N^{\mathrm{FM}} = 2 \sum_{n=0}^{N-1} \log\left(1 + \Delta t  k(t_n)\right).
\]
Using the first-order Taylor expansion, \(\log(1 + x) = x + \mathcal{O}(x^2)\), and retaining only the leading term,
\[
\log s_N^{\mathrm{FM}} = 2 \sum_{n=0}^{N-1} \left[ \Delta t  k(t_n) + \mathcal{O}\left(\frac{k(t_n)^2}{N^2}\right) \right] = 2 \sum_{n=0}^{N-1} \Delta t  k(t_n) + \mathcal{O}\left(\frac{1}{N}\right).
\]
The summation can be viewed as a Riemann sum, and subsequently approximated as an integral:
\[
\log s_N^{\mathrm{FM}} = 2 \int_0^1 k(t)  dt + \mathcal{O}\left(\frac{1}{N}\right).
\]
Since $k(t)$ is continuously differentiable in $[0,1]$, the approximation error of this integral is also $\mathcal{O}(N^{-1})$.

Using \eqref{eq:gauss-id}, we have \(k(t) = A(t)/B(t)\) and \(B'(t) = 2A(t)\), which yield,
\[
2 \int_0^1 k(t)  dt = \int_0^1 \frac{B'(t)}{B(t)}  dt = \log B(1) - \log B(0).
\]
Since \(B(0) = 1\) and \(B(1) = \sigma^2\), it follows that
\[
\log s_N^{\mathrm{FM}} = \log \sigma^2 + \mathcal{O}\left(\frac{1}{N}\right).
\]
Exponentiating both sides gives
\begin{equation}
\label{eq:sn_FM_is_ON_inv}
    s_N^{\mathrm{FM}} = \exp\left(\log \sigma^2 + \mathcal{O}\left(\frac{1}{N}\right)\right)
    = \sigma^2 \left(1 + \mathcal{O}\left(\frac{1}{N}\right)\right)
    = \sigma^2 + \mathcal{O}\left(\frac{1}{N}\right).
\end{equation}

Now consider the KL divergence between \(p_{t_N}^{\mathrm{FM}} = \mathcal{N}(\mu, s_N^{\mathrm{FM}} I_d)\) and the target \(\mathcal{N}(\mu, \sigma^2 I_d)\), given by:
\begin{equation}
    \label{eq:fm_scaling_kl}
    \mathrm{KL} = \frac{d}{2} \left( \frac{s_N^{\mathrm{FM}}}{\sigma^2} - 1 - \log\left( \frac{s_N^{\mathrm{FM}}}{\sigma^2} \right) \right).
\end{equation}
Let us denote \(\varepsilon_N \triangleq (s_N^{\mathrm{FM}} - \sigma^2)/\sigma^2\). Then from \eqref{eq:sn_FM_is_ON_inv}, $\epsilon_N = \mathcal{O}(N^{-1})$. 
Using the Taylor series expansion of $\log(1 + x)$, we get,
\[
\mathrm{KL} = \frac{d}{2} \left( \varepsilon_N - \log(1 + \varepsilon_N) \right)
= \frac{d}{2} \left( \varepsilon_N - (\varepsilon_N - \tfrac{1}{2} \varepsilon_N^2 + \mathcal{O}(\varepsilon_N^3)) \right)
= \frac{d}{4} \varepsilon_N^2 + \mathcal{O}(\varepsilon_N^3) = \mathcal{O}\left(\frac{1}{N^2}\right).
\]
This finishes the derivation of the convergence rate for FM.

{\bf Asymptotic decay rate for TM.}
The derivation for the convergence rate of TM relies on the fact that the inner ODE iterations for sampling the difference latent are FM iterations themselves.
Hence, we require an expression for how the variance contraction factor, $r_N^{\rm FM}$ from \eqref{eq:rcfm-rec} depends on the number of iterations, $N$.
This was unnecessary for proving Thm \ref{theorem:kl}, but is needed to bound the error accumulation from the inner TM iterations.

Firstly, recall from \eqref{eq:finiteS-attn} that the true covariance of the difference latent is $\operatorname{Cov} \left(\widetilde V_{t_n}\mid \widetilde X_{t_n}^{\mathrm{TM}}\right) = c_S \tau^2(t_n) I_d$, where $\tau_n^2(t) = \frac{\sigma^2}{B(t_n)}$, and the variance contraction factor $c_S \leq 1$ plays the same role as $r_n^{\rm TM}$ from \eqref{eq:rcfm-rec} in the FM analysis.
Hence, for the remainder of this proof, we bound $r_N^{\rm FM}$, while an analogous bound holds for $c_S$, with $N$ replaced by $S$.
Note that $c_S$ does not depend $n$, since once $\widetilde{X}_{t_n}^{\rm TM}$ is fixed, $c_S$ characterizes the variance contraction of the inner ODE iterations, and hence, depends only on $S$, the total number of inner ODE steps.  

The evolution of $r_n^{\rm FM}$ is specified by \eqref{eq:rcfm-rec} and \eqref{eq:wndef}.
Therefore, $r_N^{\rm FM} = \prod_{n=0}^{N-1}w_n$.
Recall the notation $a_n = (1 + \Delta t k(t_n))$, and denote $g_n \triangleq \sigma^2/B(t_n)^2$.
Then,
\begin{align}
\label{eq:log_wn}
    w_n = \frac{a_n^2}{a_n^2 + (\Delta t)^2g_n} \implies \log w_n = \log a_n^2 - \log\left(a_n^2 + (\Delta t)^2g_n\right) = -\log\left(1 + \frac{(\Delta t)^2g_n}{a_n^2}\right).
\end{align}

To expand $a_n^{-2}$, consider the Taylor series expansion, $(1+x)^{-2} = 1 - 2x + 3x^2 + {\cal O}(x^3)$ for $\lvert x \rvert < 1$.
Here, $x = \Delta t k(t_n)$ and as $k(t_n) = A(t_n)/B(t_n)$ is bounded, $\lvert \Delta t k(t_n) \rvert = \lvert k(t_n)/N \rvert < 1$ for large enough $N$.
Therefore,
\begin{align}
    a_n^{-2} &= (1 + \Delta t k(t_n))^{-2} = 1 - 2\Delta t k(t_n) + 3(\Delta t)^2k(t_n)^2 + {\cal O}(\Delta t^3) \nonumber \\
    \frac{(\Delta t)^2g_n}{a_n^2} &= (\Delta t)^2 g_n - 2(\Delta t)^3g_nk(t_n) + {\cal O}(\Delta t^4).
\end{align}
Next, since $\log (1 + x) = x - \frac{x^2}{2} + {\cal O}(x^3)$ for $\lvert x \rvert < 1$, and $(\Delta t)^2g_n/a_n^2 < 1$ for large enough $N$, using \eqref{eq:log_wn}, we get,
\begin{align}
    \log w_n = -\log\left(1 + \frac{(\Delta t)^2g_n}{a_n^2}\right) = -\frac{(\Delta t)^2g_n}{a_n^2} + {\cal O}(\Delta t^4) &= -(\Delta t)^2 g_n + 2(\Delta t)^3g_nk(t_n) + {\cal O}(\Delta t^4) \nonumber \\
    &= -\frac{\sigma^2(\Delta t)^2}{B(t_n)^2} + {\cal O}(\Delta t^3).
\end{align}

Using this with \eqref{eq:rcfm-rec} yields,
\begin{align}
\label{eq:log_rn_FM}
    \log r_N^{\rm FM}= \sum_{n = 0}^{N-1}\log w_n = -\sigma^2\sum_{n=0}^{N-1}\frac{(\Delta t)^2}{B(t_n)^2} + {\cal O}\left(\frac{1}{N^2}\right) \stackrel{\rm (i)}{=} -\frac{\sigma^2}{N}\int_0^1\frac{dt}{B(t)^2} + {\cal O}\left(\frac{1}{N^2}\right),
\end{align}
where in $\rm (i)$, we use $\Delta t = 1/N$ and approximate the summation with a finite integral for large $N$.
Since $B(t)$ is bounded in $[0,1]$, the finite integral is also bounded, and is a function of $\sigma$.
Denote $P(\sigma) \triangleq \sigma^2\int_0^1B(t)^{-2}dt$.
While $P(\sigma)$ can be evaluated exactly, it is not required for the purposes of this derivation, and can simply be treated as a constant. 
Taking exponentials in \eqref{eq:log_rn_FM}, and using $e^x = 1 + x + {\cal O}(x^2)$, this gives,
\begin{align}
\label{eq:rnFM_is_1/N2}
    r_N^{\rm FM} = {\rm exp}\left(-\left(\frac{P(\sigma)}{N}\right) + {\cal O}\left(\frac{1}{N^2}\right)\right) = 1 - \frac{P(\sigma)}{N} + {\cal O}\left(\frac{1}{N^2}\right).
\end{align}

Now that we are able to show the asymptotic dependence of $r_N^{\rm FM}$ on $N$, a similar result holds true for the inner ODE iterations in TM, and is given by
\begin{align}
\label{eq:inner_ODE_variance_contraction}
    c_S = 1 - \frac{P(\tau(t_n))}{S} + {\cal O}\left(\frac{1}{S^2}\right),
\end{align}
where given $\widetilde{X}_{t_n}^{\rm TM}$, $\tau(t_n)$ is as specified in \eqref{eq:cond-gauss}.

From \eqref{eq:rtm-rec}, the evolution of the variance contraction ratio for TM, $r_n^{\rm TM}$, is given by
\begin{align}
\label{eq:rn_TM_evolution}
    r_{n+1}^{\rm TM} = w_nr_n^{\rm TM} + (1 - w_n)c_S, \quad r_0^{\rm TM} = 1, \text{ and, } w_n = \frac{a_n^2}{a_n^2 + (\Delta t)^2\sigma^2/B(t_n)^2}.
\end{align}

Since $r_{n+1}^{\rm TM}$ is a convex combination of the previous ratio, $r_n^{\rm TM}$, and $c_S$, which is independent of $n$, the evolution for $N \geq 1$ can be expressed as,
\begin{align}
    r_n^{\rm TM} = c_S + (1 - c_S)\prod_{n=0}^{N-1}w_n.
\end{align}
From \eqref{eq:inner_ODE_variance_contraction}, we can express $c_S = 1 + \delta_S$, where $\delta_S = {\cal O}\left(S^{-1}\right)$.
Therefore, using this,
\begin{align}
    s_{N,S}^{\rm TM} = r_N^{\rm TM}B(1) = \sigma^2\left(c_S + (1-c_S)r_N^{\rm FM}\right) = \sigma^2\left(1 + \delta_S(1 - r_N^{\rm FM})\right).
\end{align}
Hence, proceeding as done after \eqref{eq:fm_scaling_kl}, the relative variance error is
\begin{align}
    \varepsilon_{N,S} \triangleq \frac{s_{N,S} - \sigma^2}{\sigma^2} = \delta_S\left(1 - r_N^{\rm FM}\right),
\end{align}
and the KL divergence is then give by,
\begin{align}
    \nonumber
    {\rm KL} &= \frac{d}{2}\left(\varepsilon_{N,S} - \log\left(1 + \varepsilon_{N,S}\right)\right) = \frac{d}{4}\varepsilon_{N,S}^2 + {\cal O}\left(\varepsilon_{N,S}^3\right) = {\cal O}\left(\delta_S^2\left(1 - r_N^{\rm FM}\right)^2\right) \\
    &= {\cal O}\left(\frac{\left(1 - r_N^{\rm FM}\right)^2}{S^2}\right) \stackrel{\rm (i)}{=} {\cal O}\left(\frac{1}{N^2S^2}\right),
\end{align}
where $\rm (i)$ utilizes the asymptotic expression for $r_n^{\rm FM}$ from \eqref{eq:rnFM_is_1/N2} to get,
\begin{align}
    \left(1 - r_N^{\rm FM}\right)^2 = \left(\frac{P(\sigma)}{N} + {\cal O}\left(\frac{1}{N^2}\right)\right)^2 = {\cal O}\left(\frac{1}{N^2}\right).
\end{align}
This completes the proof.

\end{proof}

\subsection{Proof of Proposition \ref{prop:local-unimodal-main}: Unimodal approximation of difference latent distribution}

\begin{proposition*}
{\bf [Unimodal approximation of difference latent distribution]}
\label{subsec:unimodal_approx_diff_appendix}
    Let $X_0 \sim \mathcal{N}(0, I_d)$ be sampled from the standard Gaussian distribution, and let
    \[
    X_1 \sim \sum_{j = 1}^K \pi_j \mathcal{N}(\mu_j, \sigma_j^2I_d)
    \]
    be sampled from a Gaussian mixture with $K$ components, where $\pi_j > 0, \sigma_j > 0$, and $\sum_{j=1}^K\pi_j = 1$.
    Assume that $X_0$ and $X_1$ are independent.
    For any fixed $t \in (0,1]$, define the linear interpolation $X_t = (1-t)X_0 + tX_1$, and let $B_t(j) = (1-t)^2 + t^2 \sigma_j^2$ be the variance conditioned on $X_1$ being drawn from $j^{\rm th}$ component (i.e., $Z = j$).
    Given $x \in \mathbb{R}^d$, let $k_t(x) \in {\rm argmin}_{j}\lVert x - t\mu_j\rVert$ denote the index of the nearest path mean (with ties are broken arbitrarily), and let $D_t(x) \triangleq \lVert x - t\mu_{k_t(x)}\rVert$ be the distance to it.
    Furthermore, define the associated margin,
    \begin{equation*}
        \rho_t(x) \triangleq {\rm min}_{j \neq k_t(x)} \left(\lVert x - t\mu_j \rVert - \lVert x - t\mu_{k_t(x)} \rVert\right),
    \end{equation*}
    to denote the gap between the distances to the closest and the second-closest path means.
    Then, for $V = X_1 - X_0$,
    \begin{align*}
        \lVert p(V \vert X_t = x) - p(V \vert X_t = x, Z = k_t(x)) \rVert_{\rm TV} \leq \left(\frac{1}{\pi_{k_t(x)}}-1\right)\hspace{-1mm}\left(\frac{B^{\rm max}_t}{B^{\rm min}_t}\right)^{\frac{d}{2}}\hspace{-1mm}{\rm exp}\hspace{-1mm}\left(\frac{D_t^2(x)}{2}\left(\frac{1}{B^{\rm min}_t} - \frac{1}{B^{\rm max}_t}\right)-\frac{\rho_t^2(x)}{2B^{\rm max}_t}\right),  
    \end{align*}
    where $B^{\rm min}_t \triangleq {\rm min}_j B_t(j)$ and, $B^{\rm max}_t \triangleq {\rm max}_jB_t(j)$
\end{proposition*}

\begin{proof}
    Conditioned on being sampled from the $j^{\rm th}$ Gaussian component, we have,
    \begin{align}
        X_t \vert Z = j \sim  \mathcal{N}\left(t\mu_j, B_t(j) I_d\right).
    \end{align}
    For a Gaussian mixture model, {\it posterior responsibility} refers to the posterior probability that a given data point was generated from a specific Gaussian component.
    Denote the posterior responsibilities as
    \begin{align}
        w_t(x,j) \triangleq \mathbb{P}(Z = j \vert X_t = x) = \frac{\pi_j B_t(j)^{-d/2}{\rm exp}\left(-\lVert x - t\mu_j \rVert^2/2B_t(j)\right)}{\sum_{\ell = 1}^K\pi_\ell B_t(\ell)^{-d/2}{\rm exp}\left(-\lVert x - t\mu_\ell \rVert^2/2B_t(\ell)\right)}
    \end{align}

    Marginalizing over the probability of the components, we have,
    \[
    p(V \vert X_t = x) = \sum_{j}w_t(x,j)p(V \vert X_t = x, Z = j)
    \]
    Utilizing $\sum_{\ell} w_t(x, \ell) = 1$ and triangle inequality,
    \begin{align}
    \label{eq:upper_bound_TV_1}
        \lVert p(V \vert X_t = x) - p(V \vert X_t = x, Z = k) \rVert_{\rm TV} &= \left\lVert \sum_jw_t(x,j)p(V \vert X_t = x, Z = j) - p(V \vert X_t = x, Z = k)\right\rVert_{\rm TV} \nonumber \\
        &\leq \sum_{j \neq k} w_t(x,j)\left\lVert p(V \vert X_t = x, Z = j) - p(V \vert X_t = x, Z = k) \right\rVert_{\rm TV} \nonumber \\
        &\stackrel{\rm (i)}{\leq} \sum_{j \neq k} w_t(x,j) = 1 - w_t(x,k).
    \end{align}
    Here, $\rm (i)$ follows from the fact that TV distance is always upper bounded by $1$.
    Therefore, it suffices to upper bound \eqref{eq:upper_bound_TV_1}.
    To this end, for $k \equiv k_t(x)$, define the {\it responsibility ratio} as,
    \begin{align}
    \label{eq:responsibility_ratio_definition}
        r_{j\vert k}(x) \triangleq \frac{w_t(x,j)}{w_t(x,k)} = \frac{\pi_j}{\pi_k}\left(\frac{B_t(x,k)}{B_t(x,j)}\right)^{\frac{d}{2}}{\rm exp}\left(-\frac{\lVert x - t\mu_j \rVert^2}{2B_t(x,j)} + \frac{\lVert x - t\mu_k \rVert^2}{2B_t(x,k)}\right).
    \end{align}
    We also have,
    \begin{align}
    \label{eq:upper_bound_responsibility}
        w_t(x,k) = \frac{1}{1 + \sum_{j \neq k}r_{j \vert k}(x)} \implies 1 - w_t(x,k) = \frac{\sum_{j \neq k}r_{j \vert k}(x)}{1 + \sum_{j \neq k}r_{j \vert k}(x)} \stackrel{\rm (i)}{\leq} \sum_{j \neq k} r_{j \vert k}(x),
    \end{align}
    where $\rm (i)$ follows from the fact that the denominator $1 + \sum_{j \neq k}r_{j \vert k}(x)$ always exceeds $1$.
    To upper bound $\sum_{j \neq k} r_{j \vert k}(x)$, note that the expression inside the ${\rm exp}(\cdot)$ in \eqref{eq:responsibility_ratio_definition} is,
    \begin{align}
        -\frac{\lVert x - t\mu_j \rVert^2}{2B_t(x,j)} + \frac{D_t^2(x)}{2B_t(x,k)} \leq -\frac{\lVert x - t\mu_j \rVert^2}{2B_t^{\rm max}} + \frac{D_t^2(x)}{2B_t^{\rm min}} \stackrel{\rm (i)}{\leq} -\frac{\rho_t^2(x)}{2B_t^{\rm max}} + \frac{D_t^2(x)}{2}\left(\frac{1}{B_t^{\rm min}} - \frac{1}{B_t^{\rm max}}\right),
    \end{align}
    where to get $\rm (i)$, we add and subtract $\frac{D_t^2(x)}{2B_t^{\rm max}}$ and use the fact that $\lVert x - t\mu_j \rVert \geq D_t(x) \hspace{-1mm} \implies \hspace{-1mm} \lVert x - t\mu_j \rVert^2 - D_t^2(x) \geq \rho^2_t(x)$.
    Substituting this in \eqref{eq:upper_bound_responsibility} yields,
    \begin{align}
        \sum_{j \neq k} r_{j \vert k}(x) \leq {\rm exp}\left(\frac{D_t^2(x)}{2}\left(\frac{1}{B^{\rm min}_t} - \frac{1}{B^{\rm max}_t}\right)-\frac{\rho_t^2(x)}{2B^{\rm max}_t}\right)\sum_{j \neq k}\frac{\pi_j}{\pi_k}\left(\frac{B_t(x,k)}{B_t(x,j)}\right)^{d/2}.
    \end{align}
    Upper bounding $\frac{B_t(x,k)}{B_t(x,j)} \leq \frac{B_t^{\rm max}}{B_t^{\rm min}}$ and noting that $\sum_{j \neq k}\frac{\pi_j}{\pi_k} = \frac{1}{\pi_{k_t(x)}}-1$, completes the proof.
\end{proof}

\subsection{Proof of Corollary \ref{cor:localized-bounds}: Effect of mode separation}
\label{app:effect_of_mode_separation}

\begin{corollary*}
{\bf [Effect of mode separation]}
\label{cor:localized-bounds_appendix}
Under the setting of Prop.~\ref{prop:local-unimodal-main}, and $\rho_t(x) \triangleq \min_{j \ne k_t(x)} \big(\|x - t\mu_j\| - \|x - t\mu_{k_t(x)}\|\big)$,
define minimal mean separation, $D_{\min}\triangleq \min_{j\ne k}\|\mu_j-\mu_k\|$ and the neighborhood $\mathcal{E}_t \triangleq \big\{x \in \mathbb{R}^d : D_t(x) \le \sqrt{B_t^{\min}} \big\}$. 
Assuming $B_t^{\max} = B_t^{\min}$, for any $x \in \mathcal{E}_t$, 
\[
    \big\|p(V\mid X_t=x)-p(V\mid X_t=x, Z=k_t(x))\big\|_{\mathrm{TV}}
    \leq
    \Big(\frac{1}{\pi_{k_t(x)}}-1\Big)
    \exp \left(
    2 - \frac{t^2 D_{\min}^2}{4B_t^{\max}}
    \right).
\]
\end{corollary*}

\begin{proof}
Starting from the original bound presented in Prop.~\ref{prop:local-unimodal-main}, 
\[
\lVert p(V \vert X_t = x) - p(V \vert X_t = x, Z = k_t(x)) \rVert_{\rm TV} \leq \left(\frac{1}{\pi_{k_t(x)}}-1\right)\hspace{-1mm}\left(\frac{B^{\rm max}_t}{B^{\rm min}_t}\right)^{\frac{d}{2}}\hspace{-1mm}{\rm exp}\hspace{-1mm}\left(\frac{D_t^2(x)}{2}\left(\frac{1}{B^{\rm min}_t} - \frac{1}{B^{\rm max}_t}\right)-\frac{\rho_t^2(x)}{2B^{\rm max}_t}\right).
\]

From the definition of the neighborhood, $D_t(x)$ is upper bounded by $\sqrt{B_t^{\min}}$, and the bound gives
\begin{equation}
    \label{eq:tv_bound_neighbor}
    \big\|p(V\mid X_t=x)-p(V\mid X_t=x, Z=k_t(x))\big\|_{\mathrm{TV}}
    \le 
    \Big(\frac{1}{\pi_{k_t(x)}}-1\Big)
    \Big(\frac{B_t^{\max}}{B_t^{\min}}\Big)^{ \frac{d}{2}}
    \exp \left(
    \frac{1}{2}\Big(1-\frac{B_t^{\min}}{B_t^{\max}}\Big)
    -\frac{\rho_t^2(x)}{2B_t^{\max}}
    \right).
\end{equation}

From the definition of $D_{\min}$ and $\rho_t(x)$, i.e., 
\[
\rho_t(x) \triangleq \min_{j \ne k_t(x)} \big(\|x - t\mu_j\| - \|x - t\mu_{k_t(x)}\|\big),
\quad 
D_{\min} \triangleq \min_{j \ne k}\|\mu_j - \mu_k\|,
\] 
an application of reverse triangle inequality yields for any $j \ne k_t(x)$,
\[
\|x - t\mu_j\| \ge \|t\mu_j - t\mu_{k_t(x)}\| - \|x - t\mu_{k_t(x)}\|
\ge tD_{\min} - D_t(x).
\]
This implies, $\|x - t\mu_j\| - D_t(x) \ge tD_{\min} - 2D_t(x)$, and taking the minimum over $j \ne k_t(x)$ gives
\[
\rho_t(x) \ge tD_{\min} - 2D_t(x) 
\geq tD_{\min} - 2\sqrt{B_t^{\min}},
\]
where the second inequality follows as $D_t(x)\le \sqrt{B_t^{\min}}$. 
Hence,
\[
-\frac{\rho_t^2(x)}{2B_t^{\max}}
\ \le\
-\frac{\big(tD_{\min}-2\sqrt{B_t^{\min}}\big)^2}{2B_t^{\max}}.
\]

Additionally, from the inequality $(a-b)^2 \ge \frac12 a^2 - b^2$, 
the square term in the exponential gives $(tD_{\min}-2\sqrt{B_t^{\min}})^2 \ge \frac12 t^2 D_{\min}^2 - 4 B_t^{\min}$, which gives
\[
-\frac{\big(tD_{\min}-2\sqrt{B_t^{\min}}\big)^2}{2B_t^{\max}}
\ \le\
-\frac{t^2 D_{\min}^2}{4B_t^{\max}} + \frac{2B_t^{\min}}{B_t^{\max}}.
\]

Plugging in \eqref{eq:tv_bound_neighbor}, then for any $x \in \mathcal{E}_t$,
\begin{align}
    \nonumber
    &\big\|p(V\mid X_t=x)-p(V\mid X_t=x, Z=k_t(x))\big\|_{\mathrm{TV}} \\
    \le 
    &\Big(\frac{1}{\pi_{k_t(x)}}-1\Big)
    \Big(\frac{B_t^{\max}}{B_t^{\min}}\Big)^{  d/2}
    \exp  \left(
    \frac{1}{2}\Big(1-\frac{B_t^{\min}}{B_t^{\max}}\Big)
    +\frac{2B_t^{\min}}{B_t^{\max}}
    -\frac{t^2 D_{\min}^2}{4B_t^{\max}}
    \right) \\
    \nonumber
    \approx &\Big(\frac{1}{\pi_{k_t(x)}}-1\Big)
    \exp \left(
    2 - \frac{t^2 D_{\min}^2}{4B_t^{\max}}
    \right),
\end{align}
where the second approximation holds from the assumption that $B_t^{\max} = B_t^{\min}$. 
This completes the proof.

\qedhere
\end{proof}

\subsection{Local Unimodal Approximation for Gaussian Mixture}
\label{subsec:local_unimodal_approximation}

For Gaussian mixtures, after a short period, both FM or TM iterates fall into one component's basin.
From that point onwards in time, the nearest mode dominates, and the future updates effectively follow the unimodal case. 
Consequently, we can invoke Thm. \ref{theorem:kl} for the remainder of the trajectory, and conclude that TM will outperform FM.
The following Lemma bounds the probability with which a single mode dominates the local trajectory.

\begin{lemma}
\label{lemma:high_probability_good_region}
    Fix $t \in (0,1]$ and define the {\bf good region} parameterized by $(r, \rho^*)$ as,
    \begin{equation}
    \label{eq:good_region}
        {\cal G}_t(r, \rho^*) \triangleq \left\{x : \lVert x - t\mu_{k_t(x)}\rVert \leq r \text{ and } \rho_t(x) \geq \rho^* \right\},
    \end{equation}
    where the nearest path mean, $k_t(x)$ and the margin, $\rho_t(x)$ are defined in Prop.~\ref{prop:local-unimodal-main}.
    Then,
    \begin{equation}
        \mathbb{P}\left(X_t \notin {\cal G}_t(r,\rho^*)\right) \leq {\rm exp} \left(-\frac{r^2}{2B_t^{\rm min}}\right) + {\rm exp}\left(-\frac{(tD_{min} - \rho^*)^2}{8B_t^{\rm min}}\right),
    \end{equation}
    where $D_{\min}\triangleq \min_{j\ne k}\|\mu_j-\mu_k\|$ (as defined in Cor. \ref{cor:localized-bounds}), and $B^{\rm min}_t \triangleq {\rm min}_j B_t(j)$.
\end{lemma}
\begin{proof}
    From union bound and the law of total probability, we have,
    \begin{align*}
        \mathbb{P}\left(X_t \notin {\cal G}_t(r,\rho^*)\right) \leq \sum_{j}\pi_j\mathbb{P}\left(\lVert X_t - t\mu_{k_t(x)} \rVert > r \vert k_t(x) = j\right) + \sum_j\pi_j\mathbb{P}\left(\rho_t(X_t) < \rho^* \vert k_t(x) = j\right)
    \end{align*}
    Using Gaussian concentration around its mean,
    \begin{align*}
        \mathbb{P}\left(\lVert X_t - t\mu_{k_t(x)} \rVert > r \vert k_t(x) = j\right) \leq {\rm exp} \left(-\frac{r^2}{2B_t(j)}\right).
    \end{align*}
    Also, noting the fact that $\rho_t(x) < \rho^*$ implies $\lVert x - t\mu_{k_t(x)} \rVert > (tD_{min} - \rho^*)/2$, and again using Gaussian concentration, we have,
    \begin{align*}
        \mathbb{P}\left(\rho_t(X_t) < \rho^* \vert k_t(x) = j\right) \leq {\rm exp}\left(-\frac{(tD_{min} - \rho^*)^2}{8B_t(j)}\right).
    \end{align*}
    Recalling that $\sum_j \pi_j = 1$, and $B^{\rm min}_t \triangleq {\rm min}_j B_t(j)$.
    Therefore, we get
    \begin{align*}
        \mathbb{P}\left(X_t \notin {\cal G}_t(r,\rho^*)\right) \leq {\rm exp} \left(-\frac{r^2}{2B_t^{\rm min}}\right) + {\rm exp}\left(-\frac{(tD_{min} - \rho^*)^2}{8B_t^{\rm min}}\right),
    \end{align*}
    completing the proof.
\end{proof}

Now, for any $x \in {\cal G}_{r, \rho^*}$, let $k = k_t(x)$.
For every $j \neq k$,
\begin{align}
\label{eq:upper_bound_rt}
    \frac{{\cal N}(t\mu_j, B_t(j)I_d)}{{\cal N}(t\mu_k, B_t(k)I_d)} &= \left(\frac{B_t(k)}{B_t(j)}\right)^{d/2} {\rm exp}\left(-\frac{\lVert x- t\mu_j \rVert^2}{2B_t(j)} + \frac{\lVert x- t\mu_k \rVert^2}{2B_t(k)}\right) \nonumber \\
    &\stackrel{\rm (i)}{\leq} \left(\frac{B_t(k)}{B_t(j)}\right)^{d/2} {\rm exp}\left(-\frac{\rho^{*2}}{2B_t(j)} + \frac{r^2}{2B_t(k)}\right),
\end{align}
where $\rm (i)$ follows from $\lVert x -t\mu_k \rVert \leq r$, and $\lVert x - t\mu_j \rVert \geq \rho^*$ (loose bound from margin).
Subsequently, the mixture can be written as:
\begin{align*}
    \pi_k{\cal N}(t\mu_k, B_t(k)I_d) + \sum_{j \neq k} \pi_j {\cal N}(t\mu_j, B_t(j)I_d) &= \pi_k{\cal N}(t\mu_k, B_t(k)I_d) \left(1 + \sum_{j \neq k}\frac{\pi_j {\cal N}(t\mu_j, B_t(j)I_d)}{\pi_k{\cal N}(t\mu_k, B_t(k)I_d)}\right) \nonumber \\
    &\stackrel{\rm (i)}{=} \pi_k{\cal N}(t\mu_k, B_t(k)I_d) (1 + \zeta_t(x)),
\end{align*}
where, using \eqref{eq:upper_bound_rt} we have,
\begin{align*}
    \zeta_t(x) \leq \left(\frac{1 - \pi_k}{\pi_k}\right) \left(\frac{B_t(k)}{B_t(j)}\right)^{d/2} {\rm exp}\left(-\frac{\rho^{*2}}{2B_t(j)} + \frac{r^2}{2B_t(k)}\right).
\end{align*}
Clearly, for large margin $\rho^*$ and small $r$, $\zeta_t(x)$ goes to zero, and the $k^{\rm th}$ Gaussian component dominates.

\subsection{Convergence of FM and TM iterates for a Gaussian mixture target}
For all the proofs in the section, let $p_1 = \sum_{j=1}^{K}\pi_j\phi_j$ be the Gaussian mixture density, where $\pi_j > 0$, $\sum_j \pi_j = 1$, and $\phi_j \equiv {\cal N}(\mu_j, \sigma_j^2I_d)$, and denote $D_{\min} = \min_{i \neq j}\lVert \mu_i - \mu_j \rVert$.
The majority of the proofs for Gaussian mixture will proceed component-wise, i.e., given a sample $X$, we will consider the likelihood that a particular component of the mixture was used for sampling.
For this purpose, we introduced the random variable $Z \in \{1, \ldots, K\}$ with $\mathbb{P}\left(Z = j\right)$ and conditional $X \vert Z = j \sim \phi_j$.

We prove our main result (Thm. \ref{thm:gaussian_mixture_KL_convergence}) through a series of Lemmas.
We will use the notion of {\it good region} \eqref{eq:good_region}, i.e.,
\begin{equation*}
    {\cal G}_t(r, \rho^*) \triangleq \left\{x : \lVert x - t\mu_{k_t(x)}\rVert \leq r \text{ and } \rho_t(x) \geq \rho^* \right\}.
\end{equation*}
Lem.~\ref{lemma:local_component_attraction} shows that if the FM/TM Euler steps bring any iterate close enough to a specific component, then subsequent iterates will also be close to that component.

\begin{lemma}
\label{lemma:local_component_attraction}
    {\bf (Local component attraction)}
    For any $M \in \{0, \ldots, N-1\}$ and $\beta \in \left(0,\frac{1}{2}\right)$, set $r = \beta t_MD_{\min}$ and $\rho^* = (1 - 2\beta)t_MD_{\min}$.
    Fix some $M \in \{0, \ldots, N-1\}$, and let $\sigma_{\rm max} = {\rm max}_j \sigma_j$, $B_{\rm max}(t) = (1-t)^2 + \sigma^2_{\rm max}$ and $B_* = {\max}_{m \geq M} B_{\rm max}(t_m) \leq (1 - t_M)^2 + \sigma_{\rm max}^2$.
    If at the ``hitting time'' $M$, the iterate lies in the good region for some $k$, i.e.,
    \begin{align*}
        \widetilde{X}_{t_M}^A \in \mathcal{G}_{t_M}(r, \rho^*) = \left\{x : \lVert x - t_M\mu_{k_{t_M}(x)} \rVert \leq r, \hspace{2ex} \rho_{t_M}(x) \geq \rho^*\right\} \quad \text{with } k_{t_M}\hspace{-1mm}\left(\widetilde{X}_{t_M}^A\right) = k,
    \end{align*}
    then for all $m \geq M$, we have,
    \begin{align*}
        \widetilde{X}_{t_m}^A \in \mathcal{G}_{t_m}\left(r, \rho^*\right) \quad \text{and} \quad k_{t_m}\hspace{-1mm}\left(\widetilde{X}_{t_m}^A\right) = k, \quad \text{ with probability exceeding } 1 - \delta,
    \end{align*}
    where $\delta = (N- M + 1){\rm exp}\left(-\frac{1}{2}\left(\frac{r}{\sqrt{B_*}} - \sqrt{d}\right)_+^2\right)$.
\end{lemma}

\begin{proof}
    For either algorithm $A \in \{{\rm FM, TM}\}$, conditioning on a specific component label $Z = k$, from the proof of Thm. \ref{thm:tm_le_cfm_kl}, we have,
    \begin{align}
        \mathbb{E}\left[{\widetilde X}_{t_n}^A \vert Z = k\right] = t_n \mu_k, \quad \operatorname{Cov}\left({\widetilde X}_{t_n}^A \vert Z = k\right) = s_{n,k}^AI_d, \quad \text{where } s_{n,k}^A \leq B_{t_n}(k) = (1 - t_n)^2 + t_n^2\sigma_k^2.
    \end{align}
    We assume that at the {\it hitting time} $M$, the iterate lies in the good region for some $k$, i.e.,
    \begin{align}
        \widetilde{X}_{t_M}^A \in \mathcal{G}_{t_M}(r, \rho^*) \quad \text{and } k_{t_M}(\widetilde{X}_{t_M}^A) = k.
    \end{align}

    Firstly, we show that if we can keep ${\widetilde X}_{t_m}$ inside the Euclidean ball, $\left\{x : \lVert x - t_m\mu_k\rVert \leq r\right\}$ for all future steps $m \geq M$, then it automatically implies that ${\widetilde X}_{t_m} \in \mathcal{G}_{t_m}(r, \rho^*)$ for all $m \geq M$.
    To see this, note that because $t_m$ is non-decreasing in $m$ and $\beta \in \left(0, \frac{1}{2}\right)$, for every $m \geq M$,
    \begin{align*}
        r = \beta t_M D_{\min} \leq \beta t_m D_{\min}, \quad \text{and } \rho^* = (1 - 2\beta)t_M D_{\min} \leq (1 - 2\beta)t_m D_{\min}.
    \end{align*}
    Hence, if $\lVert x - t_m\mu_k \rVert \leq r$, then for any $j \neq k$,
    \begin{align}
    \label{eq:within_ball_implies_margin_for_future}
        \lVert x - t_m\mu_j \rVert \geq t_m\lVert \mu_j - \mu_k \rVert - \lVert x - t_m\mu_k \rVert \geq t_m D_{\min} - \lVert x - t_m \mu_k\rVert \stackrel{\rm (i)}{\geq} \lVert x - t_m \mu_k\rVert + \rho^*,
    \end{align}
    where $\rm (i)$ follows because,
    \begin{align*}
        2\lVert x - t_m\mu_k \rVert + \rho^* \leq 2r + \rho^* = 2\beta t_M D_{\min} + (1 - 2\beta)t_M D_{\min} = t_M D_{\min} \leq t_m D_{\min}.
    \end{align*}

    In other words, \eqref{eq:within_ball_implies_margin_for_future} implies $\left\{ x : \lVert x - t_m\mu_k \rVert \leq r\right\} \subseteq \mathcal{G}_{t_m}\left(r, \rho^*\right) \text{ for all } m \geq M$.
    Hence, for the remainder of the proof, we only need to show that $\mathbb{P}\left(\lVert \widetilde{X}_{t_m} - t_m\mu_k \rVert\right) \leq r$ for all $m \geq M$.

    Now, note that conditioned on $Z = k$, we have $\widetilde{X}_{t_m}^A - t_m\mu_k \sim \mathcal{N}\left(0, s_{m,k}^A I_d\right)$.
    Using Gaussian concentration, we obtain the following uniform bound over $m \geq M$:
    \begin{align}
        \mathbb{P}\left(\lVert \widetilde{X}_{t_m}^A - t_m\mu_k \rVert > r \mid Z = k\right) \leq {\rm exp}\left(-\frac{1}{2}\left(\frac{r}{\sqrt{B_*}} - \sqrt{d}\right)_+^2\right).
    \end{align}
    An application of union bound over the event,
    \begin{align*}
        \mathcal{E} \triangleq \bigcap_{m = M}^N \left\{ \lVert \widetilde{X}_{t_m}^A - t_m\mu_k \rVert \leq r\right\} \implies \mathbb{P}\left(\mathbb{E}^{\rm C} \mid Z = k\right) \leq (N - M + 1){\rm exp}\left(-\frac{1}{2}\left(\frac{r}{\sqrt{B_*}} - \sqrt{d}\right)_+^2\right) \triangleq \delta.
    \end{align*}
    In other words, $\mathbb{P} \left( \forall m \geq M : \widetilde{X}_{t_m}^A \in \mathcal{G}_{t_m}(r, \rho^*) \mid Z = k\right) \geq 1 - \delta$, completing the proof.
\end{proof}

Next, the following Lemma shows that if any $x$ is within the good region for a component $k$, then with a high probability that component is dominant with respect to the posterior responsibilities.

\begin{lemma}
    \label{lemma:good_region_implies_responsibility_dominance}
    Fix $t \in [0,1]$.
    Suppose $x \in \mathcal{G}_t(r, \rho^*)$, then $w_t(x,k) \geq 1 - \epsilon$, where
    \begin{align}
        \epsilon \leq \left(\frac{B_t^{\rm max}}{B_t^{\rm min}}\right)^{d/2}\left(\frac{1 - \pi_k}{\pi_k}\right){\rm exp}\left(-\frac{(\rho^*)^2}{2B_t^{\rm max}} + \frac{r^2}{2B_t^{\rm min}}\right).
    \end{align}
\end{lemma}

\begin{proof}
    Note that,
    \begin{align}
    \label{eq:resp_at_odds}
        1 - w_t(x,k) = \frac{\sum_{j \neq k} \pi_k \phi_t(j)}{\pi_k\phi_t(k) + \sum_{j \neq k}\pi_j\phi_t(j)} \leq \sum_{j \neq k}\frac{\pi_j\phi_t(j)}{\pi_k\phi_t(k)},
    \end{align}
    where $\phi_t(j) \equiv \mathcal{N}(t\mu_j, B_t(j)I_d)$, and,
    \begin{align*}
        \frac{\phi_t(j)}{\phi_t(k)} = \left(\frac{B_t(k)}{B_t(j)}\right)^{d/2}{\rm exp}\left(-\frac{\lVert x - t\mu_j \rVert^2}{2B_t(j)} + \frac{\lVert x - t\mu_k \rVert^2}{2B_t(k)}\right) \stackrel{\rm (i)}{\leq} \left(\frac{B_t(k)}{B_t(j)}\right)^{d/2}{\rm exp}\left(-\frac{(\rho^*)^2}{2B_t^{\rm max}} + \frac{r^2}{2B_t^{\rm min}}\right),
    \end{align*}
    where $\rm (i)$ follows because $x \in \mathcal{R}_t(r, \rho^*)$ and $k_t(x) = k$.
    Summing over all $j \neq k$, eq. \eqref{eq:resp_at_odds} simplifies to,
    \begin{align*}
        1 - w_t(x,k) \leq \left(\frac{B_t^{\rm max}}{B_t^{\rm min}}\right)^{d/2}\left(\frac{1 - \pi_k}{\pi_k}\right){\rm exp}\left(-\frac{(\rho^*)^2}{2B_t^{\rm max}} + \frac{r^2}{2B_t^{\rm min}}\right).
    \end{align*}
    This completes the proof.
\end{proof}

\subsection{Proof of Thm. \ref{thm:gaussian_mixture_KL_convergence}: Mixture of Gaussians Target Distribution}
\label{subsec:gaussian_mixture_KL_convergence_appx}
\begin{theorem}
    \label{thm:gaussian_mixture_KL_convergence_appx}
    For any $M \in \{0, \ldots, N-1\}$ and $\beta \in \left(0,\frac{1}{2}\right)$, set $r = \beta t_MD_{\min}$ and $\rho^* = (1 - 2\beta)t_MD_{\min}$.
    Suppose $\widetilde{X}_{t_M}^{(\cdot)} \in \mathcal{G}_{t_M}(r, \rho^*)$, where $(\cdot)$ is either FM or TM, and $r$ and $\rho^*$.
    Then,
    \begin{align*}
        {\rm KL}(p_1^{\rm TM} \| p_1) < {\rm KL}(p_1^{\rm FM} \| p_1) - \gamma,
    \end{align*}
    where $\gamma$ can be arbitrarily close to $0$, and $p_1^{(\cdot)}$ are the marginal distributions of the final iterates.
\end{theorem}

\begin{proof}
    From Lem.~\ref{lemma:local_component_attraction}, the final iterates for FM and TM satisfy, $\widetilde{X}_1^{(\cdot)} \in \mathcal{G}_1(r, \rho^*)$ with probability exceeding $1 - \delta$.
    Subsequently, from Lem.~\ref{lemma:good_region_implies_responsibility_dominance}, with probability exceeding $1 - \delta$, the posterior responsibility of the final iterate satisfies $w_1(\widetilde{X}_1^{(\cdot)}, k) \geq 1 - \epsilon$.
    From Lem.~\ref{lemma:KL_mixture_vs_component}, for $\Delta_{\rm FM}, \Delta_{\rm TM} \in [\log(1 - \epsilon), 0]$, we have,
    \begin{align*}
        {\rm KL}(p_1^{\rm FM} \| p_1) = {\rm KL}(p_1^{\rm FM} \| \phi_k) + \Delta_{\rm FM}, \quad \text{and} \quad {\rm KL}(p_1^{\rm TM} \| p_1) = {\rm KL}(p_1^{\rm TM} \| \phi_k) + \Delta_{\rm TM}.
    \end{align*}
    From the unimodal analysis Thm. \ref{theorem:kl}, we have ${\rm KL}(p_1^{\rm TM} \| \phi_k) < {\rm KL}(p_1^{\rm FM} \| \phi_k)$.
    This implies,
    \begin{align}
        {\rm KL}(p_1^{\rm FM} \| p_1) - {\rm KL}(p_1^{\rm TM} \| p_1) > \Delta_{\rm FM} - \Delta_{TM} \triangleq \gamma.
    \end{align}
    Since $\Delta_{\rm FM}$ and $\Delta_{\rm TM}$ can be made arbitrarily close to $0$, $\gamma$ is also arbitrarily close to $0$, completing the proof.
\end{proof}

\section{PRELIMINARIES: PROBABILITY THEORY}
This section states some useful results from probability theory that are useful for proving the main results in App. \ref{app:proofs}. 
Most of them can be easily proved.

\begin{lemma}
    \label{lem:joint-gaussian-conditional-distribution}
    {\bf (Conditional distribution of jointly Gaussian vectors)}
    Let $(Y,X)$ be jointly Gaussian-distributed, and let its joint density in block-partitioned format be represented as
    \[
        \begin{bmatrix}
            Y \\
            X
        \end{bmatrix}
        \sim
        \mathcal{N}\left(
        \begin{bmatrix}
            \mu_Y \\
            \mu_X
        \end{bmatrix},
        \begin{bmatrix}
            \Sigma_{YY} & \Sigma_{YX} \\
            \Sigma_{XY} & \Sigma_{XX}
        \end{bmatrix}
        \right).
    \]
    Then, the conditional distribution $Y \vert X = x$ is also Gaussian with mean and covariance given by,
    \begin{equation*}
        \mathbb{E}[Y \vert X = x] = \mu_Y + \Sigma_{YX}\Sigma_{XX}^{-1}(x - \mu_X) \quad \text{and, } \quad \operatorname{Cov}(Y \vert X = x) = \Sigma_{YY} - \Sigma_{YX}\Sigma_{XX}^{-1}\Sigma_{XY}.
    \end{equation*}
    
\end{lemma}

\begin{proof}
    Consider the inverse Covariance matrix $J \triangleq \Sigma^{-1}$ in block format as
    \[
        J \;=\;
        \begin{bmatrix}
        J_{YY} & J_{YX} \\
        J_{XY} & J_{XX}
        \end{bmatrix}.
    \]
    Then, the joint Gaussian density can be written as,
    \[
        p(y,x) \;\propto\; 
        \exp\left(
          -\tfrac{1}{2}
          \begin{bmatrix}
            y - \mu_Y \\
            x - \mu_X
          \end{bmatrix}^{\top}
          J
          \begin{bmatrix}
            y - \mu_Y \\
            x - \mu_X
          \end{bmatrix}
        \right).
    \]
    When $x$ is fixed, the terms depending on $y$ are
    \[
        -\tfrac{1}{2}(y - \mu_Y)^{\top} J_{YY} (y - \mu_Y)
        \;-\; (y - \mu_Y)^{\top} J_{YX} (x - \mu_X)
        \;+\; \text{const}(x).
    \]
    
    Completing the square in $y$ yields,
    \[
        -\tfrac{1}{2}\,
        \bigl(y - \mu_Y + J_{YY}^{-1} J_{YX}(x - \mu_X)\bigr)^{\top}J_{YY}\bigl(y - \mu_Y + J_{YY}^{-1} J_{YX}(x - \mu_X)\bigr)
        \;+\;\text{const}(x).
    \]
    Therefore,
    \[
        Y \mid X = x \;\sim\;
        \mathcal N\!\bigl(\mu_Y - J_{YY}^{-1} J_{YX}(x - \mu_X),\; J_{YY}^{-1}\bigr).
    \]
    Moreover, using standard block inverse identities, 
    \[
        J_{YY}^{-1} \;=\; \Sigma_{YY} - \Sigma_{YX}\,\Sigma_{XX}^{-1}\,\Sigma_{XY},
        \qquad
        J_{YY}^{-1} J_{YX} \;=\; -\,\Sigma_{YX}\,\Sigma_{XX}^{-1}.
    \]
    Substituting these expressions above completes the proof.
\end{proof}

\begin{lemma}
    \label{lem:KL-divergence-Gaussian}
    {\bf (KL divergence between multivariate Gaussian distributions)}
    Let ${\rm P} \equiv \mathcal{N}(\mu_p, \Sigma_p)$ and ${\rm Q} \equiv \mathcal{N}(\mu_q, \Sigma_q)$.
    Then, the KL divergence between them is given by
    \begin{equation*}
        \mathrm{KL}\left({\rm P} \;\|\; {\rm Q} \right)
        = \frac{1}{2} \left[
        \operatorname{Tr}\left( \Sigma_q^{-1} \Sigma_p \right)
        + (\mu_q - \mu_p)^\top \Sigma_q^{-1} (\mu_q - \mu_p)
        - d
        + \log \frac{\det \Sigma_q}{\det \Sigma_p}
        \right].
    \end{equation*} 
\end{lemma}

\begin{lemma}
\label{lemma:KL_mixture_vs_component}
    {\bf (KL divergence of mixture and dominant component)} Let $p_1 = \sum_{j=1}^{K}\pi_j\phi_j$, where $\pi_j > 0$, $\sum_j \pi_j = 1$, and $\phi_j \equiv {\cal N}(\mu_j, \sigma_j^2I_d)$ be the Gaussian mixture density.
    Fix a component index $k \in \{1, \ldots, K\}$.
    For $x$ with $p_1(x) > 0$, denote
    \begin{equation}
        \label{eq:gaussian_mixture_resp}
        w(x,k) = \frac{\pi_k\phi_k(x)}{p_1(x)} \in (0,1].
    \end{equation}
    For any probability distribution $q$ on $\mathbb{R}^d$, assume that for some arbitrarily small $\epsilon \in [0,1)$, $w(x,k) \geq 1 - \epsilon$ for $q$-almost every $x$.
    In other words, the $k^{\rm th}$ component is dominant.
    Then,
    \begin{equation}
        {\rm KL}(q \| p_1) = {\rm KL}(q \| \phi_k) + \log\frac{1}{\pi_k} + \Delta, \quad \Delta \in [\log(1-\epsilon), 0].
    \end{equation}
    In order words, the additive gap from replacing the mixture $p_1$ by a single component $\phi_k$ is bounded by
    \begin{equation}
        \left\vert {\rm KL}(q \| p_1) - \left({\rm KL}(q \| \phi_k) + \log\frac{1}{\pi_k}\right) \right\vert \leq -\log(1 - \epsilon).
    \end{equation}
\end{lemma}

\begin{proof}
    Note that $w(x, k)$ denotes the posterior probability that $x$ was drawn from the $k^{\rm th}$ component of the mixture $p_1$.
    Using \eqref{eq:gaussian_mixture_resp}, we have,
    \begin{align}
        \log p_1(x) &= \log \pi_k + \log \phi_k(x) - \log w(x,k) \nonumber \\
        \implies {\rm KL}(q \| p_1) &\stackrel{\rm (i)}{=} \mathbb{E}_{X \sim q}\left[\log q(X) - \log p_1(X)\right] \nonumber \\
        &= \mathbb{E}_{X \sim q}\left[\log q(X) - \log \phi_k(X)\right] - \log \pi_k + \mathbb{E}_{X \sim q}[\log w(X,k)] \nonumber \\
        &= {\rm KL}(q \| \phi_k) + \log\frac{1}{\pi_k} + \Delta,
    \end{align}
    where $\rm (i)$ follows from the definition of KL divergence, and $\Delta$ denotes $\mathbb{E}_{X \sim q}[\log w(X,k)] \in [\log(1-\epsilon), 0]$.
    The bounds on $\Delta$ are obtained by noting that $1 - \epsilon \leq w(x,k) \leq 1$, completing the proof.
\end{proof}

\section{RELATED WORK}
\label{sec:related}
\paragraph{Diffusion and Flow Models.}
Diffusion models~\cite{Sohl-Dickstein:2025Thermo,Ho:2020DDPM,Song:2021SDE} synthesize data by inverting a forward process that progressively injects Gaussian noise. 
Training regresses targets implied by this corruption; the standard choice is noise prediction~\cite{Ho:2020DDPM,Song2020:DDIM}, with clean-data prediction~\cite{kingma2021variational} and $v$-prediction~\cite{salimans2022progressive} which balances the previous choices. 
Flow models~\cite{Lipman:2023CFM,Liu:2023RF,Albergo:2023Interpolant} instead consider a continuous-time path from a source to the target distribution and learn the instantaneous velocity field that transports samples along that path. 
Despite their different parameterization, both families learn conditional transition kernels determined by a supervising forward process~\cite{holderrieth2024generator, gao2024:twocoins}.
Transition Matching~\cite{shaul2025:transition} generalizes this perspective and can naturally be applied to different choices of parameterization.

\paragraph{Expectation and Conditional Velocity.}
\cite{Karras2022:EDM, xu2023:stable, biroli2024:dynamical} proposed and analyzed exact score function of an ideal denoiser in diffusion models. 
\cite{scarvelis2023:closed} leveraged exact score to investigate the generalization ability in diffusion models. 
In the context of flow models, \cite{ryzhakov2024:explicit} proposed to leverage the expectation velocity in closed-form to reduce the variance during training. 
Close to our work, \cite{bertrand2025:closed} analyzed expectation and conditional velocity, but focused on the identification on the sources of generalization in flow models. 
Previous work \cite{bertrand2025:closed} also presented analysis on the velocity distribution using \texttt{Two Moons} and CIFAR-10 datasets where the effects of data separation and variance are entangled. 
In this work, we present systemic analysis on controlled Gaussian settings to disentangle the effects of mean separation and target variance.

\begin{table}[h!]
    \setlength{\tabcolsep}{4pt}
    \scriptsize
    \centering
    \caption{
        \textbf{Number of Model Parameters and Computation Costs of TM and FM.} 
        Number of model parameters is reported in millions, and computation time is reported in seconds. 
    }
    \label{tab:model_param}
    \newcolumntype{M}{>{\centering\arraybackslash}m{0.12\linewidth}}
    \begin{tabular}{M |  M M M | M M M}
        \toprule
         & Backbone & Head & Ratio & $C_B$ & $C_H$ & $\kappa$ \\
         \midrule
         \multicolumn{7}{c}{Class-Conditioned Image Generation} \\
         \midrule
         FM & 231.15 & - & - & 0.00979 & - & - \\
         TM & 204.49 & 36.00 & 6.68 & 0.01120 & 0.00238 & 4.70 \\
         \midrule
         \multicolumn{7}{c}{Frame-Conditioned Video Generation} \\
         \midrule
         FM & 625.85 & - & - & 0.00962 & - & - \\
         TM & 627.08 & 9.08 & 69.06 & 0.00965 & 0.00024 & 40.08 \\
        \bottomrule
    \end{tabular}
\end{table}

\section{IMPLEMENTATION DETAILS}
\label{sec:impl_details}
\paragraph{Class-Conditioned Image Generation.}
For the image generation task, FM is implemented following the original DiT work~\citep{peebles2023:dit}, and TM is implemented from scratch following its original work~\citep{shaul2025:transition}. 
The TM flow head is a 6-layer MLP, with the backbone latent injected via AdaLN~\citep{peebles2023:dit}. 
Following the prior work MAR~\citep{li2024:MAR}, the images are encoded using a pretrained VAE~\citep{Rombach:2022LDM} into a latent representation of size \(16 \times 16 \times 16\). 

\paragraph{Frame-Conditioned Video Generation.}
For video generation, we adopt the History-Guided Diffusion framework~\citep{song2025:history}, which utilizes 3D DiT blocks. 
Each video frame is encoded by a pretrained VAE into a latent tensor of shape \(16 \times 16 \times 16\), and the temporal dimension is downsampled by a factor of $4$. 
Given the latent frames, we apply 3D RoPE positional embeddings~\citep{su2024:roformer} for positional encoding. 
Similar to the image generation task, the flow head is a 6-layer MLP with hidden dimension \(384\).

The number of model parameters and computation costs for each module are reported in Tab.~\ref{tab:model_param}. The backbone model size and computation cost \(C_B\) are kept comparable for both FM and TM, while the TM flow head remains relatively small and computationally efficient, with cost \(C_H\). 
Hence, the cost ratio between \(C_B\) and \(C_H\), $\kappa$, shows a several-fold difference, indicating that increasing \(S\) can be more efficient than increasing \(N\), particularly for small $N$, as shown in~\eqref{eq:scaling_cost}. 
For all other model hyperparameters, we follow the baseline configurations, and all models are trained using $32$ NVIDIA A100 GPUs.

\section{COMPARISON TO DIFFUSION MODELS}
\label{sec:appx_diff_comp}
Given a total number of steps $N$, the interval $[0, 1]$ is discretized as $t_n = n\Delta t$, where $\Delta t = 1/N$ and $n \triangleq \{0,\dots,N -1\}$. 
Following the standard diffusion-model convention, we denote $t_N = 1$ as the initial (noisy) timestep and $t_0 = 0$ as the terminal timestep, opposite to the convention adopted in the main paper. 
The source is the standard Gaussian $X_{t_N=1} \sim \mathcal N(0, I_d)$, and the target is $X_{t_0=0} \sim \mathcal N(\mu, \sigma^2 I_d)$.

\textit{DDPM (Ancestral Sampling)~\cite{Ho:2020DDPM}}: Diffusion model defines Markovian forward process with a predefined variance schedule $\beta_{t_n} = \beta_{\min} + t_n (\beta_{\max} - \beta_{\min})$: 
$$
p \left(X^{\mathrm{DM}}_{t_{n+1}} \mid X^{\mathrm{DM}}_{t_n}\right) = \mathcal{N} \left(\sqrt{1- \beta_{t_n}}X^{\mathrm{DM}}_{t_n}, \beta_{t_n} I_d \right), 
$$
where the $(\beta_{\min}, \beta_{\max})$ are chosen so that as \(n\to N\), \(p_{t_N} \to \mathcal N(0,I_d)\), e.g., $N=1000$. 
Under this choice of noise scheduling, DDPM samples follow a \textit{curved} probability path. 
The corresponding single-step reverse process is governed by the transition kernel:
$$
p \left(\widetilde X^{\mathrm{DM}}_{t_{n-1}} \mid  \widetilde X^{\mathrm{DM}}_{t_{n}} \right) = \mathcal{N} \left( \frac{ \widetilde X^{\mathrm{DM}}_{t_{n}} + \beta_{t_n} \nabla_x \log p_{t_n} \left( \widetilde X^{\mathrm{DM}}_{t_{n}} \right) }{ \sqrt{1-\beta_{t_n}} }, \beta_{t_n} I_d \right),
$$
where the variance is fixed \textit{heuristically} to $\beta_{t_n}$. 
However, ancestral sampling in DDPM requires a fine-grained time discretization (large $N$) in order to accurately approximate the target distribution.

\textit{DDIM (Skip-Step Sampler)~\cite{Song2020:DDIM}}: DDIM generalizes DDPM, allowing for skip-step sampling. 
Denote $\alpha_{t_n} = 1-\beta_{t_n}$, $\bar{\alpha}_{t_{n-1}:t_n} = \prod_{s=t_{n-1}}^{t_n} (\alpha_s)$, and $\bar{\alpha}_{t_n} = \prod_{s=1}^{t_n} (\alpha_s)$. 
Then DDIM sampler utilizes the following reverse transition kernel: 
$$
    p \left(\widetilde X^{\mathrm{DI}}_{t_{n-1}} | \widetilde X^{\mathrm{DI}}_{t_{n}} \right) = \mathcal{N} \left( \frac{\widetilde X^{\mathrm{DI}}_{t_{n}} + (1 - \bar{\alpha}_{t_{n-1}:t_n})\nabla_{x_{t_n}} \log p_\theta \left( \widetilde X^{\mathrm{DI}}_{t_{n}} \right)}{\sqrt{\bar{\alpha}_{t_{n-1}:t_n}}}, (\sigma_{t_n}^{\mathrm{DI}})^2 I_d \right), 
$$
where the variance follows the heuristic choice of DDPM, $\sigma_{t_n}^{\mathrm{DI}} = \sqrt{(1-\bar{\alpha}_{t_{n-1}})/(1-\bar{\alpha}_{t_n})} \sqrt{1-\bar{\alpha}_{t_{n-1}:t_n}}$.

\textit{Optimal Covariance Matching (OCM)~\cite{ou2024:ocm}}: OCM addresses the limitation of heuristic variances by explicitly modeling the optimal covariance of the posterior $p(x_{t_{n-1}} | x_{t_n})$:
$$ 
\Sigma_{t_{n-1}|t_n}^{\mathrm{opt}}(x) = ((1-\bar{\alpha}_{t_n})^2 \nabla_x^2 \log p_{t_n}(x) + (1-\bar{\alpha}_{t_n}) I_d) / \bar{\alpha}_{t_n},
$$ 
which can be used to replace the heuristic covariance of the previous samplers. 
In practice, the Hessian is intractable, and OCM uses an additional network that learns an approximation of this quantity. 
Using the optimal covariance gives the following transition kernel:
$$
p \left(\widetilde X^{\mathrm{OCM}}_{t_{n-1}} | \widetilde X^{\mathrm{OCM}}_{t_{n}} \right) = \mathcal{N} \left( \frac{\widetilde X^{\mathrm{OCM}}_{t_{n}} + (1 - \bar{\alpha}_{t_{n-1}:t_n})\nabla_{x_{t_n}} \log p_\theta \left( \widetilde X^{\mathrm{OCM}}_{t_{n}} \right)}{\sqrt{\bar{\alpha}_{t_{n-1}:t_n}}}, \Sigma_{t_{n-1}|t_n}^{\mathrm{opt}}(\widetilde X^{\mathrm{OCM}}_{t_{n}}) \right). 
$$

\section{ADDITIONAL RESULTS}
\label{sec:additional_results_2}


\begin{figure*}[ht!]
\centering
{\small
\setlength{\tabcolsep}{0.2em} 
\def\arraystretch{0.5}

\newcolumntype{C}{>{\centering\arraybackslash}m{0.48\textwidth}} 

\begin{tabularx}{\textwidth}{C C}
    \includegraphics[width=0.483\textwidth]{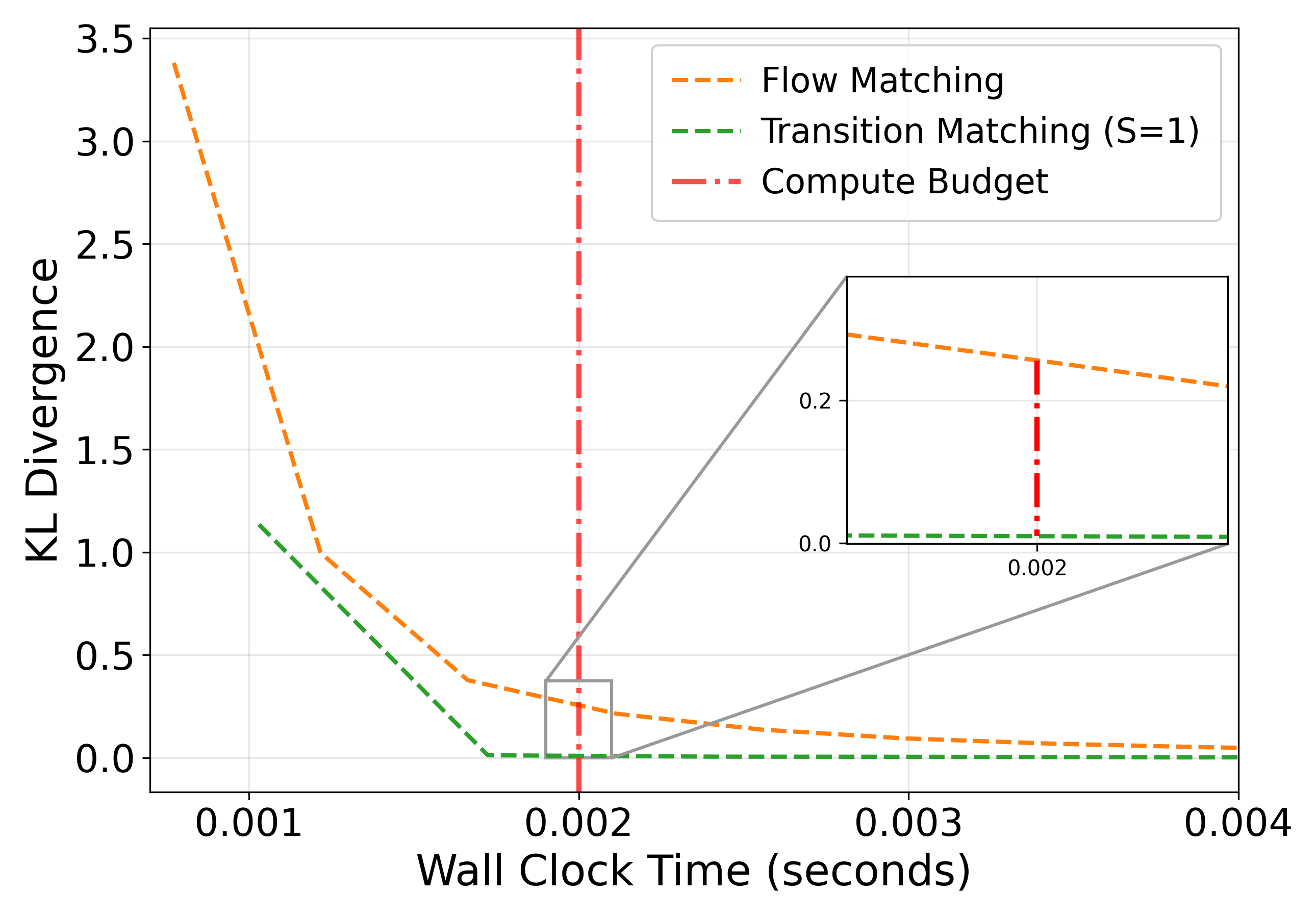} & 
    \includegraphics[width=0.48\textwidth]{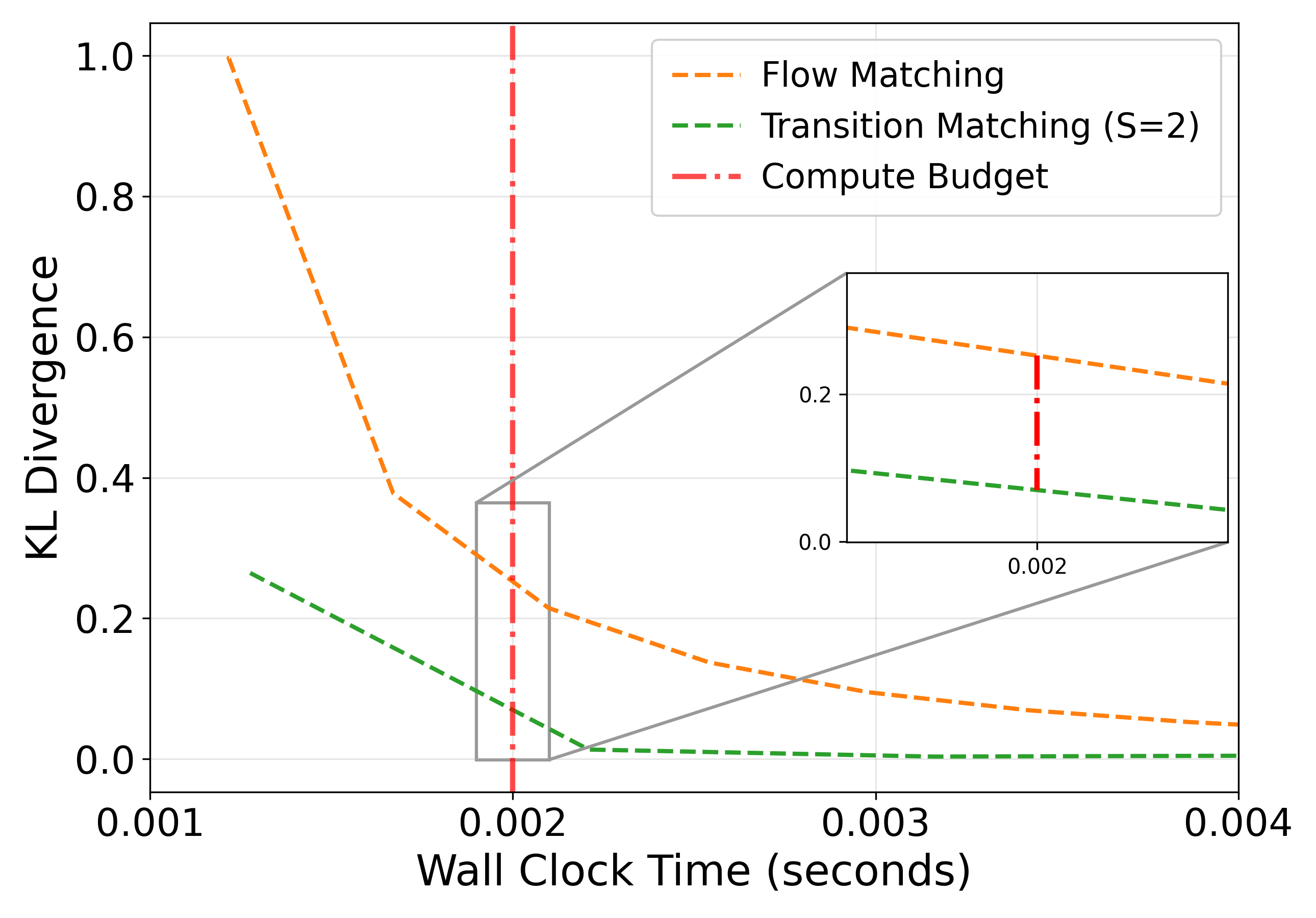} \\ 
\end{tabularx}

\caption{
    \textbf{Unimodal Gaussian KL Divergence against Wall Clock Time.}
    We compare Flow Matching (FM) and Transition Matching (TM) on unimodal Gaussian KL divergence against wall clock time. 
    For TM, we fix $S \in \{1,2\}$. The inset zooms a reference region: at matched wall-clock compute, TM achieves lower KL than FM in the low-step regime. 
}
\label{fig:unimodal_varying_steps}
}
\end{figure*}

\paragraph{Additional Results on Unimodal Gaussian.}
In this section, we extend the experiments from \S\ref{sec:gauss_target_main} and present a quantitative comparison between FM and TM under varying compute budget $N$ with fixed $S$. 
For TM, the inner ODE steps are fixed to \(S \in \{1, 2\}\), and \(N\) is chosen respectively for FM and TM to ensure comparable compute (wall clock time). 
For each task, we adopt the same evaluation setup as described in the main paper. 

The results are shown in Fig.~\ref{fig:unimodal_varying_steps}. 
While both FM and TM converge to zero KL divergence with increased compute (larger wall clock time), TM consistently outperforms FM under a fixed compute budget (red dotted line), particularly in the low-step sampling regime. 
Note that TM with a larger number of inner ODE steps (\(S = 2\)) yields lower KL divergence, reflecting the result predicted by Cor.~\ref{rem:kl_convergence} of the main paper.

\begin{figure*}[t!]
    \centering
    \includegraphics[width=\linewidth]{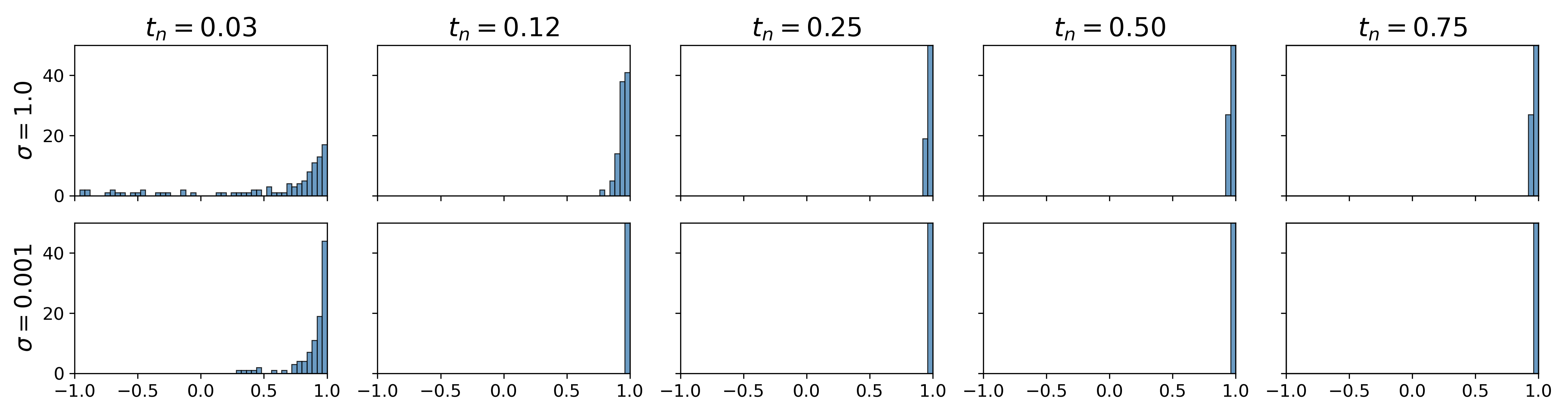}
    \caption{
        \textbf{
            Effect of Target Variance on $p(V|X)$.
        } 
        High-dimensional mixture of Gaussian with two variance settings, $\sigma \in \{ 1.0, 0.001 \}$. 
        Each panel shows the histogram of cosine similarities between difference latent samples $\widetilde V_{t_n} \sim p(V\mid X_{t_n})$ and its expectation $\mathbb{E}[V\mid X_{t_n}]$ at timestep $t_n$. 
        For $\sigma=1.0$, the difference latent samples form a unimodal distribution with non-negligible variance, whereas for $\sigma=0.001$, they concentrate near $1$, indicating collapse toward the mean. 
    }
    \label{fig:vel_cos_sim_variance}
\end{figure*}
\paragraph{Additional Results on Effect of Target Variance.}
Extending the experiment presented in \S \ref{sec:gauss_target_main}, we analyze the effect of target variance on difference latent distribution. 
Here, the target Gaussian mixture is 64-dimensional with $D_{\text{min}} = 45$. 
As shown in Fig.~\ref{fig:vel_cos_sim_variance} (row 1), when $\sigma = 1.0$, the cosine similarities cluster near \(1\) with non-negligible variance, indicating an approximation to a unimodal distribution with non-zero variance. 
Hence, following Thm.~\ref{theorem:kl}, in this regime with non-negligible variance, stochastic latent updates in TM preserve covariance and can outperform FM. 
As discussed in \S \ref{sec:gauss_target_main}, when the variance approaches zero, the difference latent distribution approximates Dirac distribution \eqref{eq:cond-gauss}. 
Fig.~\ref{fig:vel_cos_sim_variance} (row 2) shows this collapse: the cosine similarity is sharply concentrated near \(1\), indicating that the sampled difference latent is nearly identical to its expectation. Since \(\mathbb{E}[V \mid X_{t_n}]\) coincides with the velocity used in the FM update \eqref{eq:FM_expected_velocity_update}, the distinction between FM and TM vanishes.

\textbf{Comparison of FM and TM in Anisotropic Gaussian.} Let $X_0 \sim \mathcal{N}(0, I_d)$ and $X_1 \sim \mathcal{N}(\mu, \Sigma)$ be independent Gaussian vectors in $\mathbb{R}^d$, where $\Sigma = \text{diag}(\sigma_1^2, \dots, \sigma_d^2)$ is a diagonal covariance. 
Note that since the source covariance $I_d$ and the target covariance $\Sigma$ are both diagonal, and $X_0 \perp X_1$, the components of the vector $X_t = (1 - t)X_0 + tX_1$ are mutually independent for all $t \in [0, 1]$. Consequently, the $d$-dimensional problem decouples, and the proof follows a similar logic to the isotropic case shown in Thm.~\ref{theorem:kl}. 

\textit{Empirical Evaluation.} In Tab.~\ref{tab:anisotropic}, we evaluate FM and TM on a 2D anisotropic Gaussian target with diagonal covariance $\Sigma = \text{diag}(0.5, 1.5)$. We compare FM (scaling $N$) against TM (scaling $S$ with $N = 1$). As shown in the table, TM yields lower KL divergence than FM in the few-step regime, confirming its superior efficiency in Gaussian target with anisotropic covariance.

\begin{table}[h]
    \centering
    \small 
    \caption{\textbf{Quantitative Comparison of FM and TM in Anisotropic Gaussian.}}
    \begin{tabular}{lccccc}
        \toprule
        \textbf{Model} & \textbf{2} & \textbf{4} & \textbf{8} & \textbf{16} & \textbf{32} \\
        \midrule
        FM~\cite{Lipman:2023CFM} & 0.80 & 0.20 & 0.06 & 0.02 & 0.01 \\
        TM~\cite{shaul2025:transition} ($N=1$) & 0.13 & 0.06 & 0.05 & 0.03 & 0.01 \\
        \bottomrule
    \end{tabular}
    \label{tab:anisotropic}
\end{table}

\paragraph{Additional Results with Varying $N$ and $S$.}
We present additional quantitative results comparing FM and TM with varying the number of outer steps ($N$) and inner ODE steps ($S$). 
For image generation (Tab.~\ref{tab:metrics_time_n}-\ref{tab:metrics_time_s_n}), TM outperforms FM in both quality and efficiency. 
Specifically, TM ($N=8, S=8$) achieves a higher IS and lower FID than FM ($N=32$), while reducing inference latency. 
In the video generation task (Tab.~\ref{tab:video_gen_metrics_n}-\ref{tab:video_gen_metrics_s_n}), we observe a similar trend: FM ($N=16$) reaches higher FVD and FID than TM ($N=12, S=16$). 
Beyond scaling $N$, TM yields performance gains when scaling the inner steps $S$ with a fixed $N$, contributing to practical computational efficiency. 

\begin{table}[t!]
    \centering
    \caption{\textbf{Quantitative Results of FM in Class-Conditioned Image Generation}.}
    \label{tab:metrics_time_n}
    \setlength{\tabcolsep}{5pt}
    \begin{tabular}{lcccccc}
        \toprule
        \textbf{Metric} & \textbf{N=8} & \textbf{N=16} & \textbf{N=32} & \textbf{N=64} \\
        \midrule
        IS $\uparrow$ & $411.70^{\pm 1.07}$ & $432.47^{\pm 1.04}$ & $438.96^{\pm 1.56}$ & $441.11^{\pm 0.69}$ \\
        FID $\downarrow$ & $23.73^{\pm 0.08}$ & $22.13^{\pm 0.05}$ & $22.11^{\pm 0.09}$ & $21.97^{\pm 0.14}$ \\
        \midrule
        Time (s) $\downarrow$ & 0.04 & 0.08 & 0.17 & 0.32 \\
        \bottomrule
    \end{tabular}
\end{table}

\begin{table}[ht!]
    \centering
    \caption{\textbf{Quantitative Results of TM in Class-Conditioned Image Generation}.}
    \label{tab:metrics_time_s_n}
    \setlength{\tabcolsep}{5pt}
    \begin{tabular}{llcccc}
        \toprule
        \textbf{Metric} & \textbf{Step ($S$)} & \textbf{N=8} & \textbf{N=16} & \textbf{N=32} & \textbf{N=64} \\
        \midrule
        \multirow{6}{*}{IS $\uparrow$} 
        & 2 & $436.96^{\pm 0.51}$ & $441.49^{\pm 1.13}$ & $445.05^{\pm 1.71}$ & $445.54^{\pm 1.06}$ \\
        & 4 & $438.39^{\pm 0.88}$ & $446.76^{\pm 0.97}$ & $448.63^{\pm 1.07}$ & $448.09^{\pm 0.77}$ \\
        & 8 & $440.90^{\pm 0.80}$ & $449.43^{\pm 0.84}$ & $450.58^{\pm 0.88}$ & $449.74^{\pm 0.52}$ \\
        & 12 & $441.32^{\pm 0.66}$ & $450.22^{\pm 0.91}$ & $451.39^{\pm 0.84}$ & $449.80^{\pm 1.08}$ \\
        & 16 & $441.19^{\pm 0.62}$ & $450.69^{\pm 0.85}$ & $451.80^{\pm 0.93}$ & $450.14^{\pm 1.27}$ \\
        & 20 & $441.11^{\pm 0.76}$ & $450.82^{\pm 0.86}$ & $452.13^{\pm 0.92}$ & $450.27^{\pm 1.22}$ \\
        \midrule
        \multirow{6}{*}{FID $\downarrow$} 
        & 2 & $22.06^{\pm 0.11}$ & $22.07^{\pm 0.12}$ & $21.49^{\pm 0.06}$ & $21.25^{\pm 0.06}$ \\
        & 4 & $21.86^{\pm 0.10}$ & $21.79^{\pm 0.07}$ & $21.35^{\pm 0.08}$ & $21.05^{\pm 0.11}$ \\
        & 8 & $21.46^{\pm 0.08}$ & $21.67^{\pm 0.11}$ & $21.29^{\pm 0.08}$ & $20.99^{\pm 0.10}$ \\
        & 12 & $21.23^{\pm 0.09}$ & $21.58^{\pm 0.10}$ & $21.27^{\pm 0.08}$ & $20.97^{\pm 0.04}$ \\
        & 16 & $21.11^{\pm 0.08}$ & $21.54^{\pm 0.09}$ & $21.25^{\pm 0.08}$ & $20.96^{\pm 0.06}$ \\
        & 20 & $21.03^{\pm 0.08}$ & $21.52^{\pm 0.08}$ & $21.24^{\pm 0.07}$ & $20.95^{\pm 0.06}$ \\
        \midrule
        \multirow{6}{*}{Time (s) $\downarrow$} 
        & 2 & 0.08 & 0.16 & 0.31 & 0.62 \\
        & 4 & 0.10 & 0.20 & 0.39 & 0.78 \\
        & 8 & 0.14 & 0.28 & 0.55 & 1.10 \\
        & 12 & 0.18 & 0.35 & 0.71 & 1.41 \\
        & 16 & 0.22 & 0.43 & 0.87 & 1.73 \\
        & 20 & 0.26 & 0.51 & 1.02 & 2.05 \\
        \bottomrule
    \end{tabular}
\end{table}

\clearpage 
\newpage

\begin{table}[t!]
    \centering
    \caption{\textbf{Quantitative Results of FM in Frame-Conditioned Video Generation}.}
    \label{tab:video_gen_metrics_n}
    \setlength{\tabcolsep}{8pt}
    \begin{tabular}{lcccc}
        \toprule
        \textbf{Metric} & \textbf{N=8} & \textbf{N=12} & \textbf{N=16} & \textbf{N=20} \\
        \midrule
        FVD $\downarrow$ & $113.01^{\pm 0.23}$ & $95.06^{\pm 0.15}$ & $83.69^{\pm 0.12}$ & $76.40^{\pm 0.08}$ \\
        FID $\downarrow$ & $2.65^{\pm 0.03}$ & $1.95^{\pm 0.04}$ & $1.73^{\pm 0.01}$ & $1.65^{\pm 0.01}$ \\
        \midrule
        Time (s) $\downarrow$ & 0.04 & 0.06 & 0.08 & 0.10 \\
        \bottomrule
    \end{tabular}
\end{table}

\begin{table}[ht!]
    \centering
    \caption{\textbf{Quantitative Results of TM in Frame-Conditioned Video Generation}.}
    \label{tab:video_gen_metrics_s_n}
    \setlength{\tabcolsep}{5pt}
    \begin{tabular}{llcccc}
        \toprule
        \textbf{Metric} & \textbf{Step ($S$)} & \textbf{N=8} & \textbf{N=12} & \textbf{N=16} & \textbf{N=20} \\
        \midrule
        \multirow{6}{*}{FVD $\downarrow$} 
        & 2 & $134.13^{\pm 0.19}$ & $121.00^{\pm 0.29}$ & $112.15^{\pm 0.18}$ & $105.00^{\pm 0.29}$ \\
        & 4 & $108.41^{\pm 0.31}$ & $100.14^{\pm 0.40}$ & $94.39^{\pm 0.30}$ & $89.43^{\pm 0.44}$ \\
        & 8 & $92.95^{\pm 0.25}$ & $84.60^{\pm 0.38}$ & $79.60^{\pm 0.26}$ & $76.56^{\pm 0.22}$ \\
        & 12 & $89.28^{\pm 0.24}$ & $79.73^{\pm 0.31}$ & $74.41^{\pm 0.26}$ & $71.38^{\pm 0.39}$ \\
        & 16 & $88.11^{\pm 0.23}$ & $77.61^{\pm 0.27}$ & $71.94^{\pm 0.22}$ & $68.88^{\pm 0.19}$ \\
        & 20 & $87.69^{\pm 0.22}$ & $76.48^{\pm 0.26}$ & $70.56^{\pm 0.20}$ & $67.44^{\pm 0.26}$ \\
        \midrule
        \multirow{6}{*}{FID $\downarrow$} 
        & 2 & $1.82^{\pm 0.02}$ & $1.51^{\pm 0.01}$ & $1.40^{\pm 0.02}$ & $1.35^{\pm 0.02}$ \\
        & 4 & $1.66^{\pm 0.03}$ & $1.40^{\pm 0.01}$ & $1.33^{\pm 0.02}$ & $1.30^{\pm 0.03}$ \\
        & 8 & $1.72^{\pm 0.03}$ & $1.42^{\pm 0.02}$ & $1.34^{\pm 0.03}$ & $1.31^{\pm 0.03}$ \\
        & 12 & $1.80^{\pm 0.04}$ & $1.46^{\pm 0.02}$ & $1.36^{\pm 0.03}$ & $1.33^{\pm 0.02}$ \\
        & 16 & $1.86^{\pm 0.04}$ & $1.49^{\pm 0.02}$ & $1.38^{\pm 0.02}$ & $1.35^{\pm 0.03}$ \\
        & 20 & $1.90^{\pm 0.05}$ & $1.51^{\pm 0.02}$ & $1.39^{\pm 0.02}$ & $1.36^{\pm 0.02}$ \\
        \midrule
        \multirow{6}{*}{Time (s) $\downarrow$} 
        & 2 & 0.04 & 0.06 & 0.08 & 0.11 \\
        & 4 & 0.04 & 0.07 & 0.09 & 0.11 \\
        & 8 & 0.05 & 0.07 & 0.10 & 0.12 \\
        & 12 & 0.05 & 0.08 & 0.10 & 0.13 \\
        & 16 & 0.05 & 0.08 & 0.11 & 0.14 \\
        & 20 & 0.06 & 0.09 & 0.12 & 0.15 \\
        \bottomrule
    \end{tabular}
\end{table}

\paragraph{Class-Conditioned Image Generation Qualitative Results.}
Qualitative results of class-conditioned image generation in \S \ref{subsec:image_gen} are presented in Fig.~\ref{fig:image_quali}. 
For each image, we use the same class label for both FM and TM.
Additionally, we set $N=64$ for FM and $N=16$, $S=8$ for TM. 
Note that in this setting, TM requires less compute (latency) than FM, as reported in \S \ref{subsec:image_gen}. 
Despite the lower compute, TM (bottom) generates images with intricate details, such as the decorations of the throne and the lighting of the fountain.

\paragraph{Frame-Conditioned Video Generation Qualitative Results.}
Qualitative results of frame-conditioned video generation in \S \ref{subsec:video_gen} are presented in Fig.~\ref{fig:video_qualitative}. 
The leftmost column shows the conditioning frame used for video generation, and the two groups on the right show three frames generated by FM and TM, respectively. 
Videos produced by FM often exhibit artifacts, including missing content from the conditioning frame (\eg~the presenter hand in row 1 and the right leg of the baby in row 2), whereas TM preserves previous frame content. 
Additionally, TM is also more robust to semantic drift where FM often fails (\eg~the wall in row 3 and the person’s face in row 4). Finally, FM frequently yields near-stationary motion (\eg~the legs in row 4).

\begin{center}
    \hfill \break
    \textbf{Qualitative results are presented in the following pages.}
\end{center}


\begin{table}[ht]
\centering
\begin{tabularx}{\textwidth}{X}
  \centering
  \includegraphics[width=\linewidth]{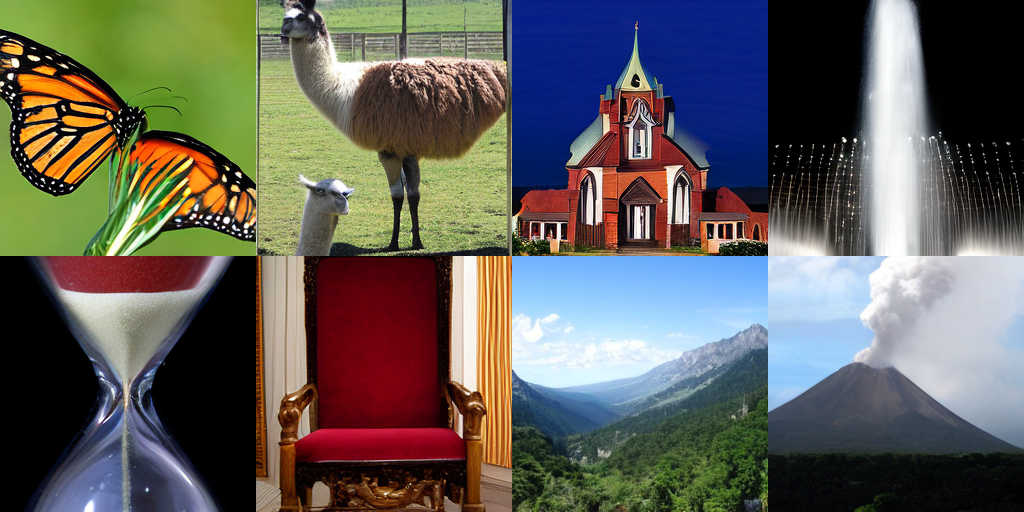}\\
  \large (a) FM~\cite{Lipman:2023CFM} \\
  \vspace{4ex}
  \includegraphics[width=\linewidth]{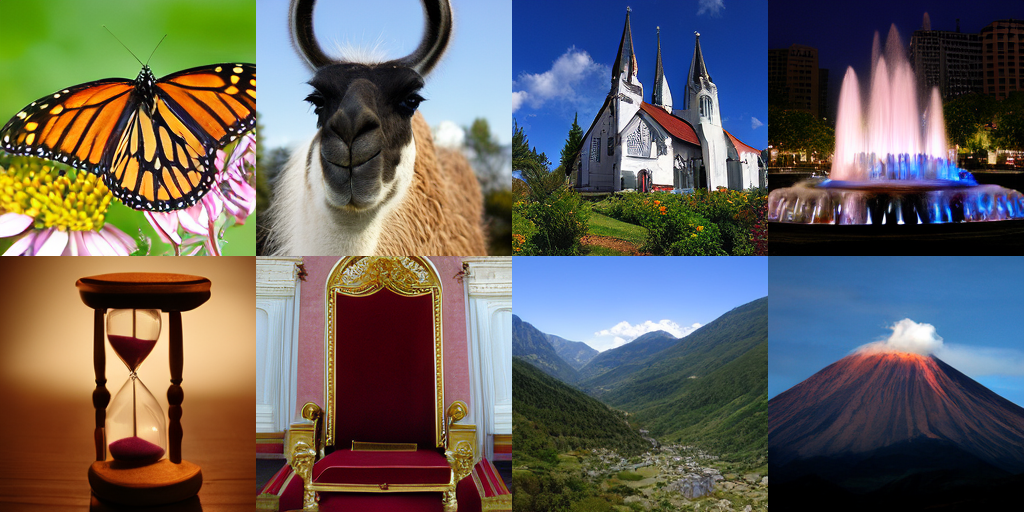}\\
  \large (b) TM~\cite{shaul2025:transition} \\
\end{tabularx}
\captionof{figure}{\textbf{Class-Conditioned Image Generation Qualitative Results.}
}
\label{fig:image_quali}
\end{table}

\clearpage
\newpage

\begin{figure*}[h!]
\centering
{\small
\setlength{\tabcolsep}{0.2em} 
\def\arraystretch{0.5}

\newcolumntype{C}{>{\centering\arraybackslash}m{0.14\textwidth}} 
\newcolumntype{Z}{>{\centering\arraybackslash}m{0.42\textwidth}} 

\begin{tabularx}{\textwidth}{C | Z | Z}
    \toprule
    \makecell{Cond. Frame} & \makecell{FM \cite{song2025:history}} & \makecell{TM \cite{shaul2025:transition}} \\
    \midrule

    \includegraphics[width=0.14\textwidth]{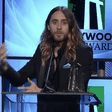} & 
    \includegraphics[width=0.42\textwidth]{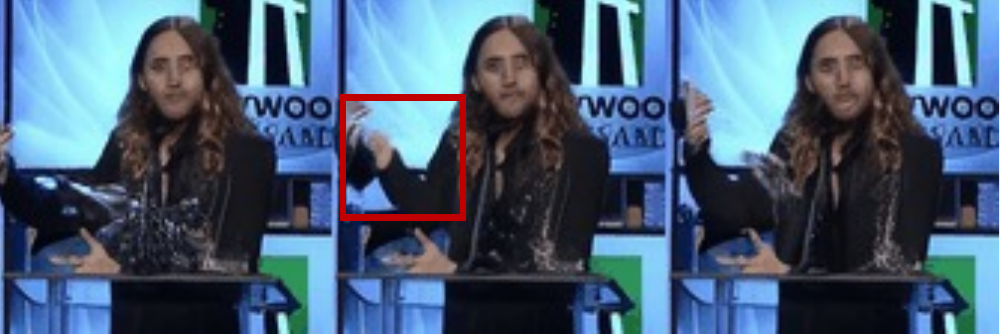} & 
    \includegraphics[width=0.42\textwidth]{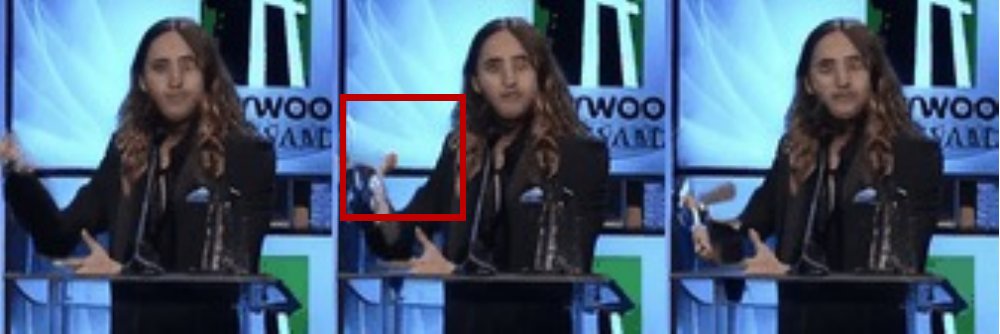} \\

    \includegraphics[width=0.14\textwidth]{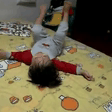} & 
    \includegraphics[width=0.42\textwidth]{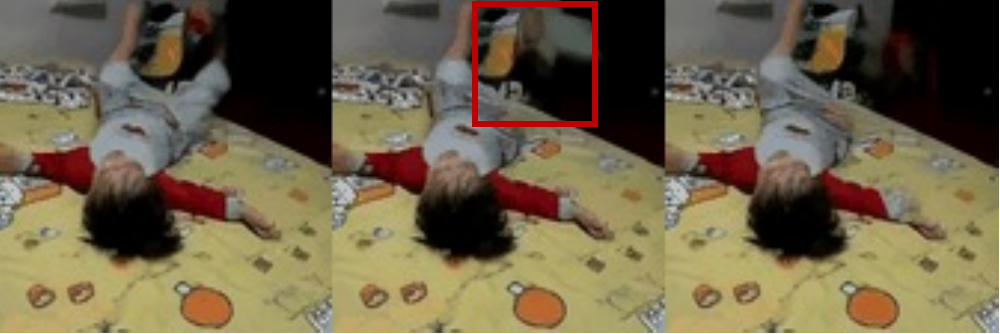} & 
    \includegraphics[width=0.42\textwidth]{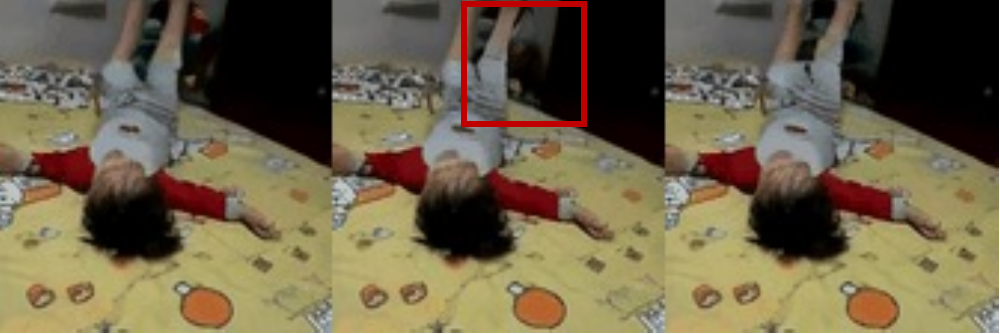} \\

    \includegraphics[width=0.14\textwidth]{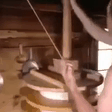} & 
    \includegraphics[width=0.42\textwidth]{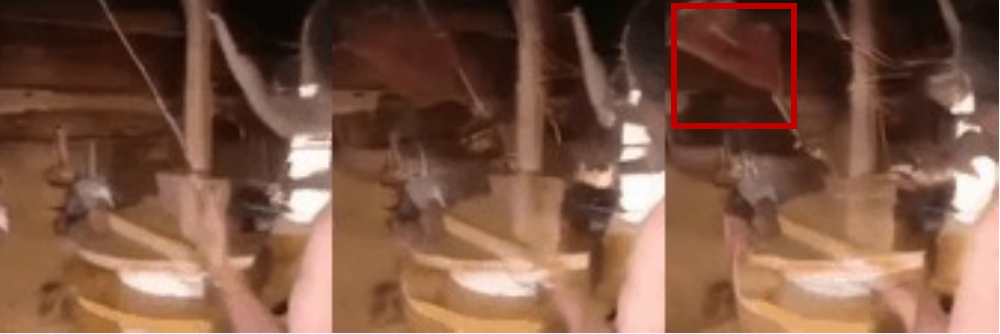} & 
    \includegraphics[width=0.42\textwidth]{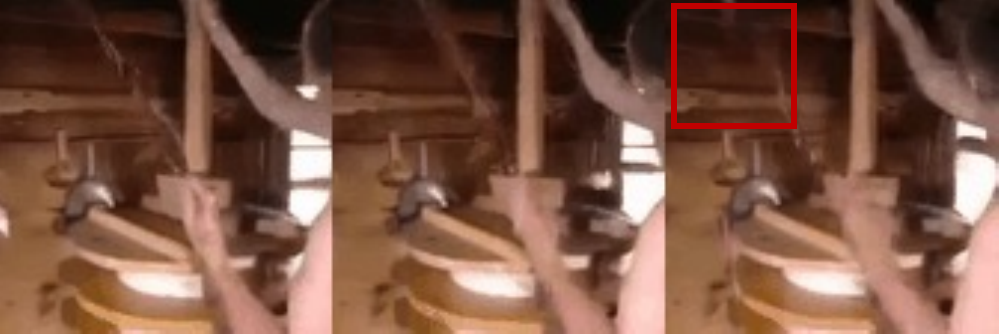} \\

    \includegraphics[width=0.14\textwidth]{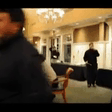} & 
    \includegraphics[width=0.42\textwidth]{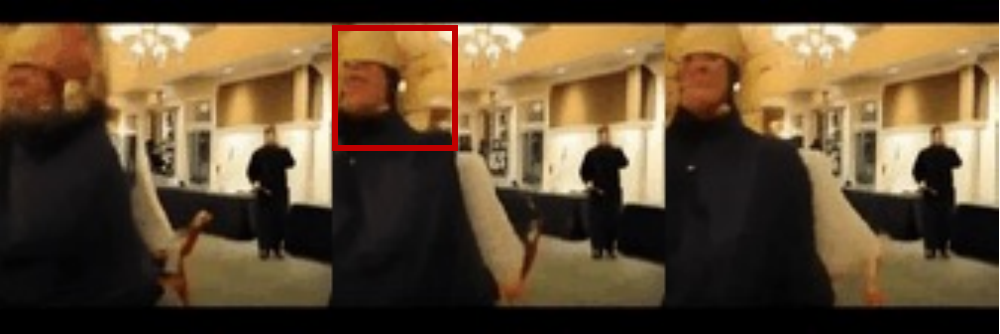} & 
    \includegraphics[width=0.42\textwidth]{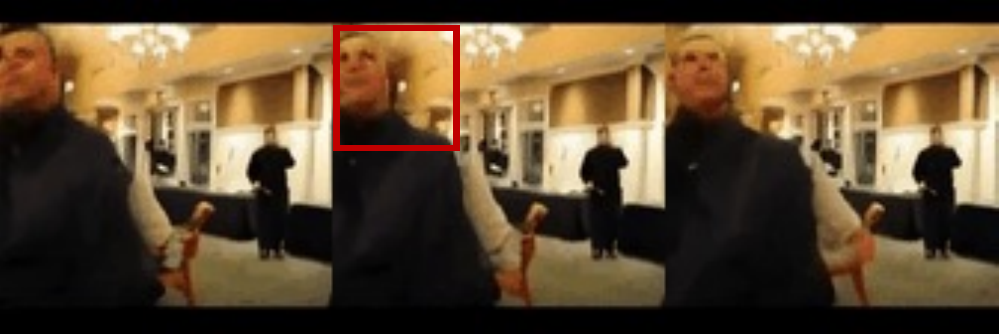} \\

    \includegraphics[width=0.14\textwidth]{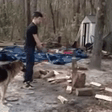} & 
    \includegraphics[width=0.42\textwidth]{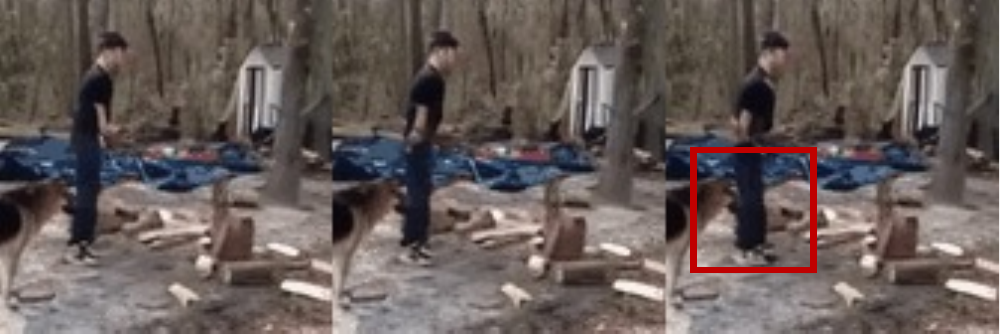} & 
    \includegraphics[width=0.42\textwidth]{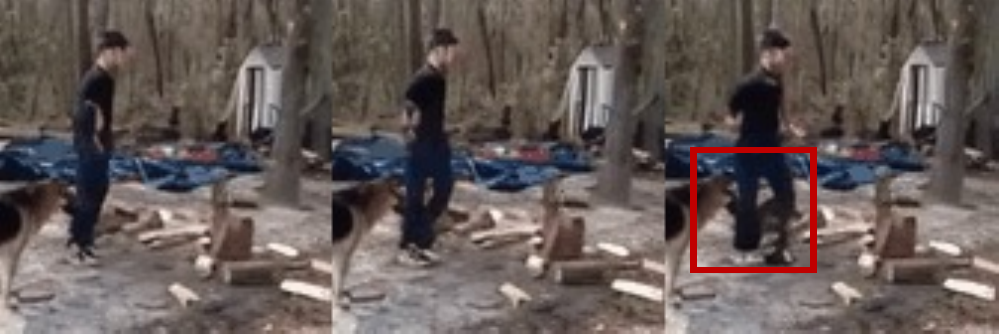} \\

    \bottomrule
\end{tabularx}

\caption{\textbf{Frame-Conditioned Video Generation Results.}}
\label{fig:video_qualitative}
}
\end{figure*}



    

    

    

\fi

\end{document}